\documentclass[lettersize,journal]{IEEEtran}
\PassOptionsToPackage{square, numbers}{natbib}
\usepackage{microtype}
\usepackage{graphicx}
\usepackage{subfigure}
\usepackage{booktabs}
\usepackage{hyperref}
\usepackage{diagbox}

\usepackage{amsmath}
\usepackage{amsfonts}
\usepackage{amssymb}
\usepackage{amsthm}

\usepackage[capitalize,noabbrev]{cleveref}
\usepackage{graphicx}
\usepackage{array}
\usepackage{multirow}
\usepackage{bbm}
\usepackage{bm}
\usepackage{enumitem}
\usepackage{makecell}

\usepackage{caption}

\usepackage[square,numbers]{natbib}

\usepackage[createShortEnv]{proof-at-the-end}

\theoremstyle{plain}
\newtheorem{theorem}{Theorem}[section]
\newtheorem{lemma}[theorem]{Lemma}

\def\BibTeX{{\rm B\kern-.05em{\sc i\kern-.025em b}\kern-.08em
    T\kern-.1667em\lower.7ex\hbox{E}\kern-.125emX}}
\usepackage{balance}
\begin{document}
\title{Understanding Normalization in Contrastive Representation Learning and Out-of-Distribution Detection}
\author{Tai Le-Gia, Jaehyun Ahn
\thanks{Tai Le-Gia and Jaehyun Ahn are with the Department of Mathematics, Chungnam National University, Daejeon, South Korea (Email: vipgiatailhp@gmail.com; jhahn@cnu.ac.kr).}}

\markboth{IEEE TRANSACTIONS ON NEURAL NETWORKS AND LEARNING SYSTEMS, VOL. XX, NO. XX, 2024 (UNDER REVIEW)}{Tai Le-Gia, Jaehyun Ahn, Understanding Normalization in Contrastive Representation Learning and Out-of-Distribution Detection}

\maketitle
\begin{abstract}
Contrastive representation learning has emerged as an outstanding approach for anomaly detection (AD). In this work, we explore the $\ell_2$-norm of contrastive features and its applications in out-of-distribution detection. We propose a simple method based on contrastive learning, which incorporates out-of-distribution data by discriminating against normal samples in the contrastive layer space. Our approach can be applied flexibly as an outlier exposure (OE) approach, where the out-of-distribution data is a huge collective of random images, or as a fully self-supervised learning approach, where the out-of-distribution data is self-generated by applying distribution-shifting transformations. The ability to incorporate additional out-of-distribution samples enables a feasible solution for datasets where AD methods based on contrastive learning generally underperform, such as aerial images or microscopy images. Furthermore, the high-quality features learned through contrastive learning consistently enhance performance in OE scenarios, even when the available out-of-distribution dataset is not suitable. Our extensive experiments demonstrate the superiority and robustness of our proposed method under various scenarios, including unimodal and multimodal settings, with various image datasets. The reproducible code is available at \url{https://github.com/nguoikhongnao/real_OECL}.
\end{abstract}


\section{Introduction}
\noindent
Out-of-distribution (OOD) detection, or anomaly detection (AD), is the task of distinguishing between normal and out-of-distribution (abnormal) data, with the general assumption that access to abnormal data is prohibited. Therefore, OOD detection methods \cite{nips-golan, nips-hendrycks, nips-csi} are typically carried out in an unsupervised manner where only normal data is available. In this regard, approaches based on self-supervised learning or pre-trained models, which are capable of extracting deep features without explicitly requiring labeled data, are commonly utilized to solve OOD detection problems.

However, the presumption that access to non-nominal data is forbidden is not highly restricted. Especially in image OOD detection problems, one can gather randomly image datasets that are likely not-normal on the internet. The utilization of such data is called Outlier Exposure \cite{iclr-oe}. Several top-performing AD methods, including unsupervised \cite{tmlr-oe}, self-supervised \cite{nips-hendrycks}, and pre-trained \cite{cvpr-panda, pmlr-deecke} approaches, utilize tens of thousands of OE samples to enhance their performance and achieve state-of-the-art detection accuracy. Nevertheless, it is important to note that Outlier Exposure (OE) also has its limitations, especially when the normal data is highly diverse \cite{tmlr-oe}, or when the OE datasets are not suitable, as demonstrated in our experiments.

One of the most effective OOD detection techniques, CSI \cite{nips-csi} is constructed by combining a special form of contrastive learning, SimCLR \cite{nips-simclr}, with a geometric OOD detection approach \cite{nips-hendrycks}. SimCLR encourages augmented views of a sample to attract each other while repelling them to a negative corpus consisting of views of other samples. CSI suggests that the addition of sufficiently distorted augmentations to the negative corpus improves OOD detection performance. Simultaneously, the network has to classify each sample’s type of transformation, as in  \cite{nips-hendrycks}. For a test sample, the anomaly score is determined by both the trained network's confidence in correctly predicting the sample's transformations and the sample's similarity to its nearest neighbor in the feature space. However, similar to geometric approaches  \cite{nips-golan, nips-hendrycks}, CSI performance depends on the relationship between normal data and shifted transformations (e.g., symmetry images and rotations).                                  

\begin{figure}
\vskip -0.2in
     \centering
         \includegraphics[width=0.5\textwidth, page=1]{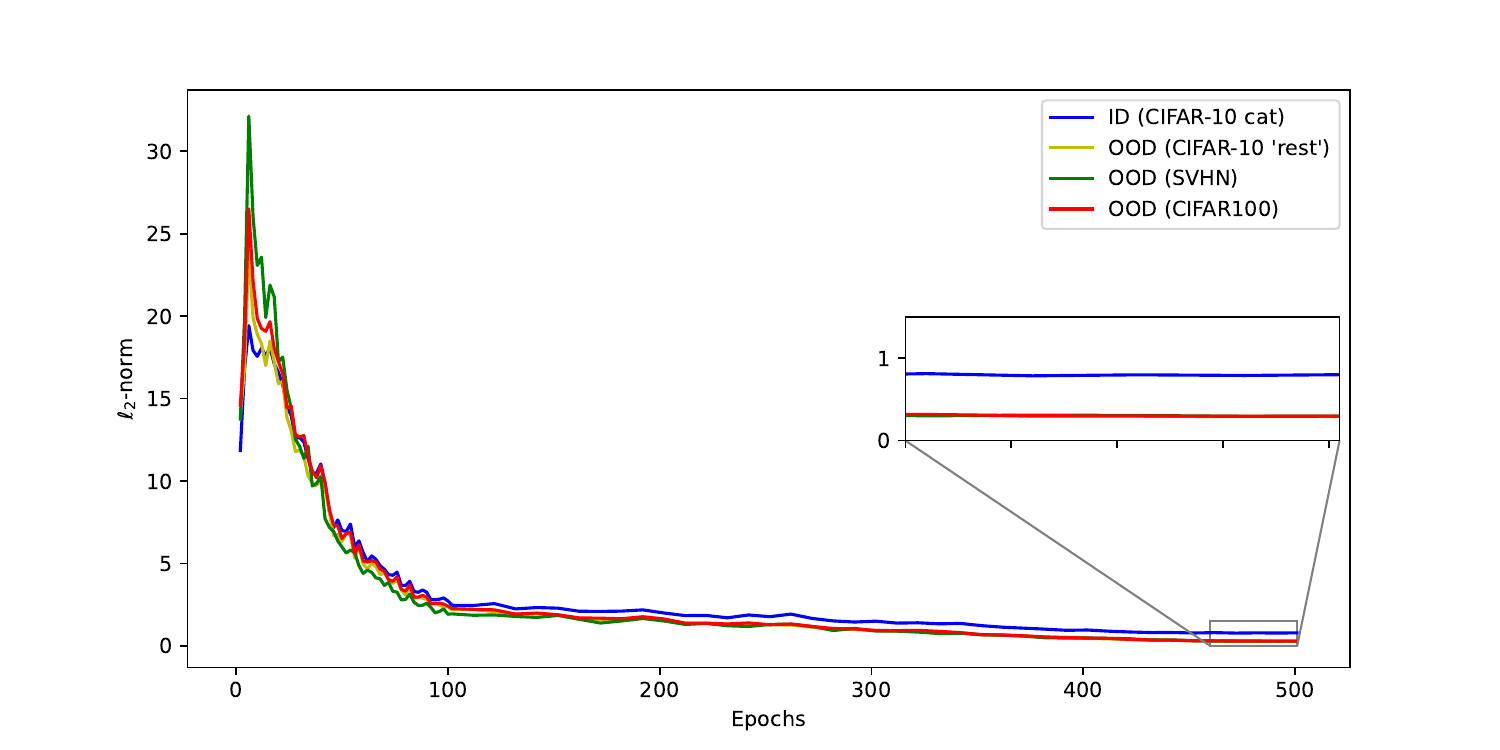}
        \caption{The $\ell_2$-norm of contrastive features for both normal and OOD samples during optimization of $\mathcal{L}_{\text{contrastive}}$ using only normal samples. Following a few epochs of rapid increase, the $\ell_2$-norm for both normal and OOD samples gradually decreases. Notably, as training progresses, the $\ell_2$-norm of the training samples becomes larger than that of the OOD samples.}
        \label{fig:1}
        \vskip 0.2in
\end{figure}

Inspired by the strengths and weaknesses of both approaches, we propose 
\emph{Outlier Exposure framework for Contrastive Learning} (OECL), an OE extension for contrastive learning OOD detection methods. By manipulating the $\ell_2$-norm of samples on the contrastive space, we found a simple way to integrate OOD samples into SimCLR-based methods. The high quality deep representations obtained by contrastive learning and the ``anomalousness" from OOD samples allow our proposed method to address the limitations of the both original approaches. To validate our claims, we conducted extensive experiments in various environments, including the common unimodal setting, the newly challenging multimodal setting.
Our main contributions are summarized as follows:
\begin{enumerate}
\item We propose a simple yet effective and robust method, OECL, to incorporate out-of-distribution data to contrastive learning OOD methods. The out-of-distribution data can be selected from an external out-of-distribution dataset or self-generated by sufficiently distorted transformations.
\item We provide insight into the $\ell_2$-normalization in contrastive learning and its application in OOD detection.
\item We empirically study the \emph{diminishing effect} of OE-based models, where ``far" OE datasets might have negative impacts to AD performance.
\end{enumerate}

The remaining sections of the paper are organized as follows. Section \ref{related-works} introduces previous works related to ours. Section \ref{section-simclr} presents SimCLR and theoretical results regarding the $\ell_2$-norm of contrastive features, as well as our proposed OECL. Extensive experiments and analysis are provided in Section \ref{experiments}, where we also study the diminishing effect. Section \ref{applicability} discusses the applicability of OECL and other OE-based methods. We conclude the paper in \cref{conclusion}.

\section{Related work} \label{related-works}
\noindent
\textbf{OOD detection.} A growing area of research uses self-supervision \cite{nips-simclr, supclr, rotations} for OOD detection. They train a network with a pre-defined task on the training set and utilize the rich features to improve OOD detection performances. These approaches have demonstrated remarkable performance in OOD detection, pushing the boundaries of what is achievable in this field \cite{nips-golan, nips-hendrycks, NEURIPS2019_6c4bb406, nips-csi, sohn2021learning}.

More recently, transfer learning-based methods have revolutionized OOD detection~\cite{dn2, pmlr-deecke,  cvpr-panda, tmlr-oe, transformly, mirzaei2022fake, musc}. These approaches involve fine-tuning supervised classifiers that were originally trained on large datasets \cite{tmlr-oe, clip}. This fine-tuning process adapts these classifiers to OOD detection tasks and has been shown to outperform the self-supervised methods on common benchmarks. However, the generalization of transfer learning-based methods to specialize data remain a concern.

Another research field in OOD detection is about incorporating outliers to improve the performance of unsupervised OOD detection methods \cite{iclr-oe, Ruff2020Deep, 9578875, tmlr-oe,  schluter2022natural, liu2023simplenet}. For instance, \cite{iclr-oe} uses OE to improve their self-supervised method by forcing the network to predict the uniform distribution for OE samples. \cite{efficientad} uses OE as a penalty to hinder the student model from generalizing its imitation of the teacher model to out-of-distribution images.

\noindent \textbf{Non-standard image OOD detection.}
Recently, there has been a growing interest in ``non-standard'' image OOD detection. Unlike the common CIFAR-10 or ImageNet, these non-standard datasets have unique characteristics that challenge OOD detection methods. One such challenge is presented by datasets where normal and abnormal samples are semantically identical but differ in their local appearances. Datasets with such a characteristic are primarily objects in defect detection research~\cite{recall, efficientad, liu2023simplenet}. Another category of non-standard datasets arises from the distinctive physical properties of training images, such as rotation-invariant images. In these cases, recent unsupervised methods \cite{nips-golan, nips-csi} have faced limitations as shown in our experiments. Datasets belong to this 'distinctive physical' category and standard images are our main focus in this paper.

\noindent \textbf{Embedding norm of normalized soft-max.} It has been observed that a normalized softmax model trained on normal data tends to yield larger feature norms over normal samples compared to OOD instances, especially in face recognition learning \cite{normface, magface, face}. NAE \cite{normface} uses $\ell_2$-norm of features as the binary person/background classification confidence. MagFace \cite{magface} assumes a positive correlation between the norm and quality of features, and use the embedding norm to penalize low-quality face samples and reward high-quality ones. In the context of general anomaly detection, CSI \cite{nips-csi} suggests using product of $\ell_2$-norm of contrastive features along with cosine similarity as the anomaly detection score.

\section{Contrastive Learning and Anomaly Detection } \label{section-simclr}
\noindent
Suppose we are given an unlabeled dataset $\mathcal{D} = \left\lbrace x_i \right\rbrace_{i=1}^{N}$ sampled from a distribution $p_{\text{data}}(x)$ over $\mathbb{R}^n$. The goal of anomaly detection is to construct a detector that identifies whether $x$ is sampled from $p_{\text{data}}$ or not. However, directly modeling an accurate model of $p_{\text{data}}$ is impractical or infeasible in most cases. To address this, many existing approaches to anomaly detection instead rely on defining a scoring function $s(x)$, with a high value indicating that $x$ is likely to belong to the in-distribution.

\subsection{Normalization in Contrastive Learning} \label{contrastive learning}
\noindent
\textbf{Contrastive learning framework.}
 Contrastive learning aims to learn an encoder by maximizing agreement between positive pairs. Empirically, given a set of pre-determined transformations $T$, positive pairs are obtained by applying randomly two transformations to the same sample, e.g., two affine transformations of the same picture $x$ \cite{nips-maximizing-views, nips-simclr, cpvr-momentum}. These positive samples are also regarded as a surrogate class of $x$ \cite{nips-dosovitskiy}. In this paper, we investigate the specific and widely popular form of contrastive loss where the training encoder $z(x) \triangleq f(x)/\lVert f(x) \rVert:  \mathbb{R}^n \to \mathcal{S}^{m-1}$ maps data to a hypersphere of dimension $m-1$. This loss has been outstandingly successful in anomaly detection literature \cite{nips-csi, iclr-ssd, nips-step},
\begin{align} \label{loss:simclr}
\mathcal{L}_{\text{contrastive}}&(z; \tau, M) \triangleq \\ \underset{\substack{(x, y) \sim p_{\text{pos}} \\ \left\lbrace x_i^- \right\rbrace_{i=1}^{M} \overset{\text{i.i.d}}{\sim} p_{\text{data}}}}{\mathbb{E}} &\left[-\log \frac{e^{z(x)^tz(y)/\tau}}{e^{z(x)^tz(y)} + \sum_i e^{z(x_i^-)^tz(y)/\tau}} \right], \end{align}
where $\tau$ is a temperature hyperparameter, $M$ is a fixed number of negative samples in a mini-batch and $p_{\text{pos}}$ is the distribution of positive pairs over $\mathbb{R}^n \times \mathbb{R}^n$.
This loss can be decoupled into \emph{alignment loss} and \emph{uniformity loss} as in \cite{icml-wang}:

\begin{align}
\mathcal{L}_{\text{align}} &= \underset{(x, y) \sim p_{\text{pos}}}{\mathbb{E}} \left[-z(x)^tz(y)/\tau\right] \label{eq:align}\\ 
\mathcal{L}_{\text{uniform}} &= \nonumber \\
 \underset{\substack{(x, y) \sim p_{\text{pos}} \\ \left\lbrace x_i^- \right\rbrace_{i=1}^{M} \overset{\text{i.i.d}}{\sim} p_{\text{data}}}}{\mathbb{E}}& \left[\log \left(e^{z(x)^tz(y)} + \sum_i e^{z(x_i^-)^tz(y)/\tau}\right)\right]\label{eq:uniform}
\end{align}

The first term $\mathcal{L}_{\text{align}}$ encourages samples from the same surrogate class be mapped to nearby features. Meanwhile, $\mathcal{L}_{\text{uniform}}$ maximizes averages distances between all samples, leading to features vectors roughly uniformly distributed on the unit hypersphere $\mathcal{S}^{m-1}$ \cite{icml-wang}.

\subsection{The $\ell_2$-norm of contrastive features}
\noindent
In the field of anomaly detection, it is observed that a normalized soft-max model trained on the normal data exhibits larger values of feature norm over normal samples than the OOD instances \cite{normface, face, magface, openset}. Particularly, CSI \cite{nips-csi} suspects that ``increasing the norm maybe an easier way to maximize cosine similarity between two vectors: instead of directly reducing the feature distance of two augmented samples, one can also increase the overall norm of the features to reduce the relative distance of two samples''. Our study sheds further light on the behavior of the $\ell_2$-norm: contrastive learning gradually reduces the $\ell_2$-norm of contrastive features, while the alignment loss compels a significantly larger $\ell_2$-norm for the contrastive features of training samples.

\noindent
\textbf{Gradual reduction of the $\ell_2$-norm of contrastive features.} This is a phenomenon we observed in many settings using contrastive loss functions based on cosine similarity. As illustrated in \cref{fig:1}, following a few epochs of rapid increase, the $\ell_2$-norm of both normal and OOD samples gradually decreases. We suspect that this phenomenon is a natural outcome of employing cosine similarity along the interaction between $\mathcal{L}_{\text{align}}$ and $\mathcal{L}_{\text{uniform}}$. We provide further detailed analysis and experiments about this phenomenon in \cref{appendix:red}.

\noindent
\textbf{A larger $\ell_2$-norm of normal contrastive features.} As training progresses, the $\ell_2$-norm of the normal samples becomes larger than that of the OOD samples. To understand this behavior, let's assume that the surrogate classes $\left\lbrace f(t(x))\right\rbrace_{t \sim T}$ are normally distributed. Since the cosine between two vectors is invariant under rotations, we can assume without loss of generality that $\mathbb{E}_{t \sim T}f(t(x)) = \left(\mu, 0, \ldots, 0 \right)$. The following theorem provides a clear illustration of the relationship between contrastive features and alignment in contrastive learning.

\begin{theoremE}[][end, restate] \label{thm:1}
Given two vectors $\bm{X} = \left(X_1, X_2, \ldots, X_n\right)^t$ and $\bm{Y} = \left(Y_1, Y_2, \ldots, Y_n\right)^t$ are i.i.d $\mathcal{N}(\boldsymbol\mu, \boldsymbol\Sigma)$ where $\boldsymbol\mu = \left(\mu, 0, \ldots, 0 \right)^t,$  $\mu\geq 0$ and 
$\bm\Sigma = 
\begin{bmatrix}
\sigma& \bm{0} \\
\bm{0} & \Sigma_V
\end{bmatrix}$. Then,
\begin{enumerate}
\item $\mathbb{E}\left\langle \frac{\bm{X}}{\lVert \bm{X} \rVert},  \frac{\bm{Y}}{\lVert \bm{Y} \rVert} \right\rangle$ is an increasing function of $\mu$
\item The following inequalities hold for all $\epsilon > 0$
\begin{align}
\label{thm:1:1} \mathbb{E}&\left\langle \frac{\bm{X}}{\lVert \bm{X} \rVert},  \frac{\bm{Y}}{\lVert \bm{Y} \rVert} \right\rangle \leq \left(\Phi\left(\frac{\mu}{\sigma}\right) - \Phi\left(-\frac{\mu}{\sigma}\right)\right)^2,  \\
\label{thm:1:2}\mathbb{E}&\left\langle \frac{\bm{X}}{\lVert \bm{X} \rVert},  \frac{\bm{Y}}{\lVert \bm{Y} \rVert} \right\rangle \geq \nonumber \\ &\frac{1}{1+\frac{\sigma_V^2}{\epsilon^2\mu^2}} \left(\Phi\left(\frac{(1-\epsilon)\mu}{\sigma}\right) \right.  - \left. \Phi\left(\frac{-(1+\epsilon)\mu}{\sigma}\right) \right)^2, \end{align} 

where $\sigma_V^2 = \text{tr}\left(\Sigma_V\right)$ and $\Phi$ is the cumulative distribution function of $\mathcal{N}(0,1)$.
\end{enumerate}
\end{theoremE}

\begin{proofEnd}
Since $EX_i = 0$ for $i \geq 2$ and they are jointly normal distribution, we have
\begin{align*}
\mathbb{E} \frac{X_i}{\sqrt{X_1^2+X_2^2+\ldots +X_n^2}} = 0 ~ \forall i \in \left\lbrace2, \dots, n \right\rbrace.
\end{align*}
Hence,
\begin{align*}
\mathbb{E}\left\langle \frac{\bm{X}}{\lVert \bm{X} \rVert},  \frac{\bm{Y}}{\lVert \bm{Y} \rVert} \right\rangle & = \mathbb{E} \frac{X_1Y_1}{\lVert \bm{X} \rVert \lVert \bm{Y} \rVert} + \sum_{i=2}^{n} \mathbb{E} \frac{X_iY_i}{\lVert \bm{X} \rVert \lVert \bm{Y} \rVert}
= \left(\mathbb{E}\left[ \frac{X_1}{\lVert \bm{X} \rVert} \right]\right)^2
\end{align*}
Let denote $U= X_1$, $V = \left(X_2, \ldots, X_n\right)^t$ and $Z = \lVert V \rVert_2^2 = \sum_{i=2}^{n} X_i^2$. The expectation of cosine similarity between two vectors $\bm{X}$ and $\bm{Y}$ becomes

\begin{align*}
 \left(\frac{1}{\sigma\sqrt{2\pi}} \int_{-\infty}^{\infty} \frac{u}{\sqrt{u^2+z}}e^{-(u-\mu)^2/2\sigma^2}du \int_{0}^{\infty} f_Z(z)dz  \right)^2,
\end{align*}
where $f_Z$ are probability density function of $Z$. 

\noindent
\begin{enumerate}
\item Let $$A = \frac{1}{\sigma\sqrt{2\pi}} \int_{-\infty}^{\infty} \frac{u}{\sqrt{u^2+z}}e^{-(u-\mu)/2\sigma^2}du \int_{0}^{\infty} f_Z(z)dz.$$
By the change of variable theorem, replacing $u = u + \mu$ give
\begin{align*}
A =  \frac{1}{\sigma\sqrt{2\pi}} \int_{-\infty}^{\infty} \frac{u+\mu}{\sqrt{(u+\mu)^2+z}}e^{-u^2/2\sigma^2}du \int_{0}^{\infty} f_Z(z)dz
\end{align*}
Get derivative,
$$A'_\mu = \frac{1}{\sigma\sqrt{2\pi}} \int_{-\infty}^{\infty} \frac{z}{\sqrt{\left((u+\mu)^2+z\right)^3}}e^{-u^2/2\sigma^2}du \int_{0}^{\infty} f_Z(z)dz > 0$$

\item For \eqref{thm:1:1}, using \cref{lemma:1} we have 
$f(z) = \int_{-\infty}^{\infty} \frac{u}{\sqrt{u^2+z}}e^{-(u-\mu)/2\sigma^2}du$ is an decreasing function on $\left[0, \infty\right)$, thus $f(z) \leq f(0) ~\forall z \geq 0$. This gives us
\begin{align*}
A = \frac{1}{\sigma\sqrt{2\pi}} & \int_{-\infty}^{\infty} \frac{u}{\sqrt{u^2+z}}e^{-(u-\mu)^2/2\sigma^2}du \int_{0}^{\infty} f_Z(z)dz \\
&\leq \left(\frac{1}{\sigma\sqrt{2\pi}} \int_{-\infty}^{\infty} \frac{u}{\lvert u \rvert}e^{-(u-\mu)^2/2\sigma^2}du\right) \left( \int_{0}^{\infty} f_Z(z)dz\right) \\
& = \frac{1}{\sigma\sqrt{2\pi}} \left( \int_{0}^{\infty}e^{-(u-\mu)^2/2\sigma^2}du - \int_{-\infty}^{0}e^{-(u-\mu)/2\sigma^2}du \right)\\
& =  \frac{1}{\sqrt{2\pi}} \left( \int_{-\mu/\sigma}^{\infty}e^{-u^2/2}du - \int_{-\infty}^{-\mu/\sigma}e^{-u^2/2}du\right) \\
& = \Phi\left(\frac{\mu}{\sigma}\right) - \Phi\left(-\frac{\mu}{\sigma}\right)
\end{align*}
For \eqref{thm:1:2}, we have $f(z)$ is convex by \cref{lemma:1}. Jensen's inequality gives us
\begin{align*}
A = \frac{1}{\sigma\sqrt{2\pi}} \mathbb{E}_{Z}f(Z) \geq \frac{1}{\sigma\sqrt{2\pi}}f(\mathbb{E}_Z(Z)).
\end{align*}
Since $\mathbb{E}Z = \mathbb{E}\left[\sum_{i=2}^{n} X_i^2 \right]= \sum_{i=2}^{n}\mathbb{E} X_i^2 = \text{tr}\left(\Sigma_V\right) = \sigma_V^2$, thus
\begin{align}  \label{thm1:eq2}
A &\geq  \frac{1}{\sigma\sqrt{2\pi}}\int_{-\infty}^{\infty} \frac{u}{\sqrt{u^2+\sigma_V^2}}e^{-(u-\mu)^2/2\sigma^2}du \nonumber\\
&= \frac{1}{\sigma\sqrt{2\pi}}\int_{0}^{\infty} \frac{u}{\sqrt{u^2+\sigma_V^2}}\left(e^{-(u-\mu)^2/2\sigma^2} - e^{-(u+\mu)^2/2\sigma^2} \right)du \nonumber \\
&> \frac{1}{\sigma\sqrt{2\pi}}\int_{\epsilon\mu}^{\infty} \frac{u}{\sqrt{u^2+\sigma_V^2}}\left(e^{-(u-\mu)^2/2\sigma^2} - e^{-(u+\mu)^2/2\sigma^2} \right)du ~\forall \epsilon>0 \nonumber \\
& \geq \frac{1}{\sigma\sqrt{2\pi}} \left(\frac{\epsilon\mu}{\sqrt{(\epsilon\mu)^2 + \sigma_V^2}}\int_{\epsilon\mu}^{\infty}  \left(e^{-(u-\mu)^2/2\sigma^2} - e^{-(u+\mu)^2/2\sigma^2} \right)du\right)\\
&= \frac{1}{\sqrt{1+\frac{\sigma_V^2}{\epsilon^2\mu^2}}} \frac{1}{\sqrt{2\pi}}\left(\int_{(\epsilon-1)\mu/\sigma}^{\infty}e^{-u^2/2}du - \int_{(\epsilon+1)\mu/\sigma}^{\infty}e^{-u^2/2}du\right)\nonumber\\
&= \frac{1}{\sqrt{1+\frac{\sigma_V^2}{\epsilon^2\mu^2}}} \left( \Phi\left(\frac{(1-\epsilon)\mu}{\sigma}\right) - \Phi\left(-\frac{(1+\epsilon)\mu}{\sigma}\right) \right) \nonumber
\end{align}
The \cref{thm1:eq2} is due to the fact that $\dfrac{u}{\sqrt{u^2+\sigma_V^2}}$ is an increasing function in $u$.
\end{enumerate}
\end{proofEnd}

\Cref{thm:1} emphasizes the significance of both ratios $\mu/\sigma$ and $\mu/\sigma_V$ in the minimization of $\mathcal{L}_{\text{align}}$. The inequality \eqref{thm:1:1} demonstrates that without a sufficiently large ratio $\mu/\sigma$, the cosine similarity between positive pairs cannot be maximized effectively, even when the variance of the contrastive feature projection onto the hyperspace perpendicular to $\mathbb{E}\left(f_t(x)\right)$ is small $\left(\Sigma_V \sim 0\right)$. Moreover, as $\mu \to \infty$ for fixed $\sigma$ and $\sigma_V$, the right-hand side of \eqref{thm:1:2} converges to $1$. This means that optimizing $\mu$ is an effective way to maximize the cosine similarity of the augmentations, as illustrated in \cref{fig:2}. The above observations indicate that, in the context of contrastive learning, the encoder undergoes a learning process that encourages a sufficiently large norm of the contrastive feature of training samples, thereby facilitating alignment. 

\begin{figure*}[t]
\vskip -0.2in
     \centering
     \subfigure[$\mu/\sigma_v.$]{\includegraphics[width=0.4\textwidth, page=1]{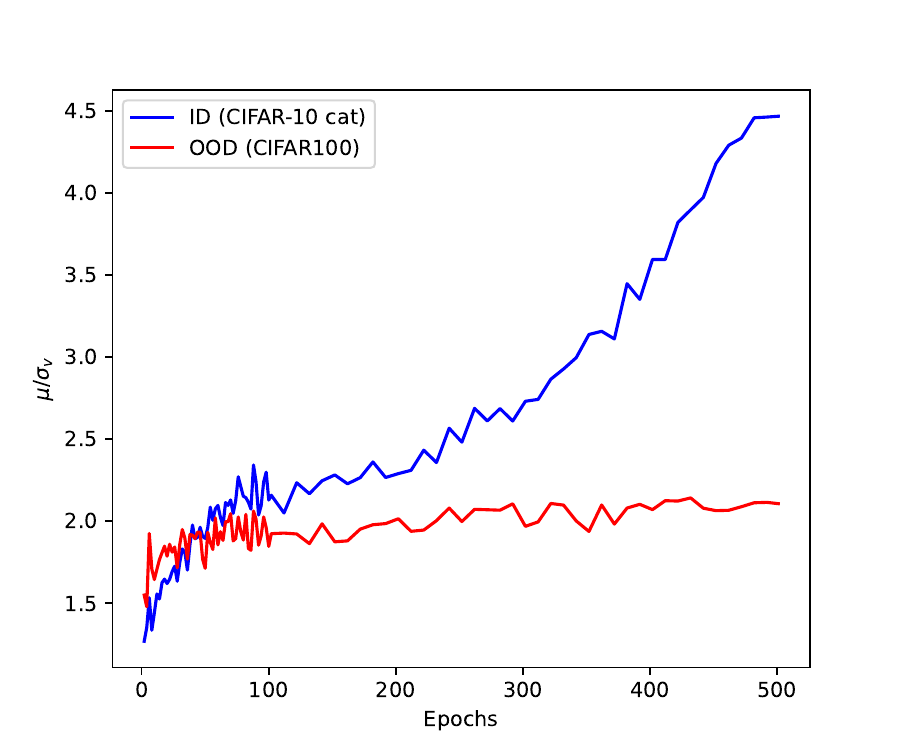}}
     \subfigure[$\sigma_v.$]{\includegraphics[width=0.4\textwidth, page=1]{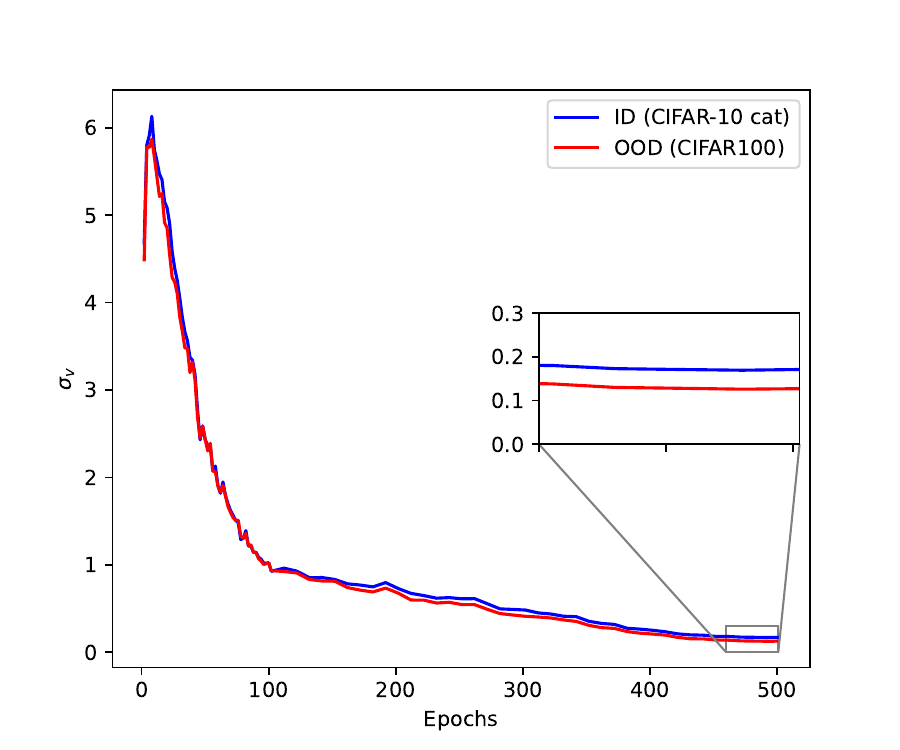}}
        \caption{Averages of $\mu/\sigma_v$ and $\sigma_v$ for both normal and OOD samples during the optimization of $\mathcal{L_{\text{contrastive}}}$ using only normal samples. While there is only a minimal difference in $\sigma_v$ between normal and OOD samples, the ratio $\mu/\sigma_v$ for normal samples becomes significantly larger as the training progresses.}
        \label{fig:2}
\vskip 0.2in
\end{figure*}

\subsection{The $\ell_2$-norm of contrastive features: more than an anomaly score}
\noindent
As discussed in \Cref{contrastive learning}, the normalization in contrastive learning inherently promotes a large norm for the contrastive features over training samples rather than samples from an out-of-distribution dataset; it naturally creates a separation between in-distribution data and out-of-distribution data on the contrastive feature space. Exploiting this characteristic, CSI uses the $\ell_2$-norm of contrastive features as a powerful tool for detecting anomalies, which has been experimentally shown to be remarkably effective \cite{nips-csi}. Beyond a tool used in inference time, we propose to utilize the $\ell_2$-norm as a component in the encoder training process.

\noindent
\textbf{Outlier exposure for contrastive learning.} Our key finding is that we can further increase this separation by simply forcing the encoder to produce \emph{zero}-values contrastive features for many augmentations of out-of-distribution samples, thus improving the anomaly detection capabilities. In this paper, we consider a specific out-of-distribution dataset, referred to as \emph{outlier exposure} dataset, denoted as $\mathcal{D}_{\text{oe}}$. We define the general form of our proposed \emph{outlier exposure contrastive learning} loss as follows:
\begin{align} \label{loss:1}
\mathcal{L}_{\text{oecl}} \triangleq \mathcal{L}_{\text{contrastive}}(z; \tau, M) + \alpha \mathbb{E}_{x \in \mathcal{D}_{\text{oe}}, t\in T_{\text{oe}}}\lVert f(t(x)) \rVert_2,
\end{align}
where $\alpha$ is a balancing hyper-parameter, and $T_{\text{oe}}$ is a set of transformations for OE augmentations. Typically, $T_{\text{oe}}$ is set equal to the set of augmentations $T$. $\mathcal{D}_{\text{oe}}$ can either consist of randomly collected images in terms of outlier exposure \cite{iclr-oe, tmlr-oe}, or be generated through distributional-shifting transformations as employed in CSI \cite{nips-csi}. This versatility allows our proposed approach to be interpreted as an outlier exposure extension for contrastive learning or a fully self-supervised generalization of CSI and SimCLR. Specifically, in the later setting, we consider a family of augmentations $\mathcal{S}_{\text{oe}}$, which we call \emph{outlier exposure transformations}. These transformations satisfy that for any $S \in \mathcal{S}_{\text{oe}}$, $S(x)$ can be regarded as an outlier with respect to any data joining $\mathcal{L}_{\text{contrastive}}$. It is worth noting that if $\mathcal{L}_{\text{contrastive}}$ is CSI, then $\mathcal{S}_{\text{oe}}$ does not contain any elements from the distribution-shifting transformations $\mathcal{S}$ of CSI, i.e., $\mathcal{S}_{\text{oe}} \cap \mathcal{S} = \varnothing$. The \emph{self-outlier-exposure contrastive learning} loss is given by,
\begin{align} \label{loss:2}
\mathcal{L}_{\text{{oecl}}} = \mathcal{L}_{\text{contrastive}}(z; \tau, M) + \alpha&\mathbb{E}_{x\in \mathcal{S}_{\text{oe}}\left(\mathcal{D}\right), t\in T_{\text{oe}}}\lVert f(t(x)) \rVert_2, 
\end{align}
where $\mathcal{S}_{\text{oe}}\left(\mathcal{D}\right) \triangleq \left\lbrace S(x) \right\rbrace_{S\in \mathcal{S}_{\text{oe}}, x \in \mathcal{D}}$.

\noindent
\textbf{Score functions for detecting anomalies.} Upon the contrastive feature $f(\cdot)$ learned by our proposed training objective, we define several score functions for deciding whether a given sample $x$ is normal or not. All of our score functions follow the conventional philosophy that in-distribution samples have higher scores. The most obvious choices are $\ell_2$-norm of the contrastive feature and its ensemble version over random augmentations.
\begin{align} 
\text{Non-ensemble}&:&s_{\ell_2} &\triangleq \lVert f(x) \rVert_2^2, \label{score:0} \\
\text{Ensemble}&:& s_{\mu} &\triangleq  \lVert \mathbb{E}_{t\in T} f(t(x)) \rVert_2^2. \label{score:1} \\
\nonumber
\end{align}
More results on anomaly score functions are in \cref{appendix:score}.
\section{Main Experiments} \label{experiments}
\noindent
In this section, we present our results for common unimodal and multimodal benchmarks, with a strong focus on the OE extension setting.
\subsection{Implementation Details}
\noindent
We train our models using the objective function \eqref{loss:1}, where $\mathcal{L}_{\text{contrastive}}$ is SimCLR. We adopt ResNet-18 \cite{resnet} as the main architecture, and we pre-train our models for 50 epochs using SimCLR. As for data augmentations $T$ and $T_{\text{oe}}$, we follow the ones used by CSI \cite{nips-csi}: namely, we apply random crop, horizontal flip, color jitter, and grayscale. If not mention otherwise, we detect OOD samples using the ensemble score function $s_{\mu}$ \eqref{score:1} with 32 augmentations. 

In addition to the outlier-exposure extension, for unimodal benchmark, we also report the results using the fully self-supervised approach, where we train our models using the objective function \eqref{loss:2} with CSI as $\mathcal{L}_{\text{contrastive}}$ and rotations as shifting transformations. Specifically, we choose $\mathcal{S} = \left\lbrace \mathbb{I}, R_{90} \right\rbrace$ and $\mathcal{S}_{\text{oe}} = \left\lbrace R_{180}, R_{270}\right\rbrace$, where $R_{\text{a}}$ is the rotation of $\text{a}^\circ$ degree. Complete experimental details are available in the \cref{appendix:exp}.

\subsection{Datasets}
\noindent
We evaluate our models using the standard CIFAR-10 \cite{cifar10} and ImageNet-30 \cite{imagenet, nips-hendrycks} datasets. We further conduct our experiments with additional datasets: the aerial dataset DIOR \cite{dior} and the microscopy dataset Raabin-WBC \cite{raabin}. Both of them are symmetry datasets, which are obviously challenging for rotation-based methods like CSI \cite{nips-csi}, GT \cite{nips-hendrycks} or our self-OECL. It is important to clarify that, in this study, the term ``non-standard'' datasets refers to datasets from domains beyond conventional life photography.

For OE datasets, we use 80 Million Tiny Images (80MTI) \cite{tiny80m} for the CIFAR-10 dataset and ImageNet-22k with ImageNet-1k removed for the other datasets. This follows the experimental setup in \cite{nips-hendrycks} and \cite{tmlr-oe}. Please refer to the \cref{appendix:exp} for more details.

\subsection{Benchmarks}
\noindent
We conduct our experiments in two main benchmarks: (1) unimodal and (2) multimodal benchmarks. 

\noindent
\textbf{Unimodal benchmark.}
Generally, unimodal is the setting where there is one semantic class in the normal data. Within this study, if not mentioned otherwise, we mostly use the widely popular one vs. rest benchmark \cite{dsvdd, nips-golan, iclr-oe, Ruff2020Deep, nips-csi, sohn2021learning, pmlr-deecke, cvpr-panda, transformly}. This benchmark is constructed based on classification datasets (e.g., CIFAR-10), where ``one" class (e.g., ``cat") is considered normal and the ``rest" classes (e.g., ``dog", ``automobile", \ldots) are considered anomalous during the test phase. Unless otherwise specified, we use all available classes as our one vs. rest classes.

\noindent
\textbf{Multimodal benchmark.} We consider a recently and more challenging multimodal benchmark called leave-one-class-out \cite{ Ahmed_Courville_2020, tmlr-oe, transformly}. In contrast to the one vs. rest benchmark, the roles are reversed, i.e., the ``rest'' classes (e.g., ``dog'', ``automobile'', \ldots) are designated as normal and the ``one'' class (e.g., ``cat'') is abnormal. The presence of multiple semantically different classes as a normal class in this benchmark makes the distribution of normal samples multimodal \cite{tmlr-oe}, hence the models are required to learn high-quality deep representations to effectively distinguish between normal and abnormal samples. For all datasets, all available classes are employed as our leave-one-class-out classes. However, due to computational constraints, for ImageNet-30, we randomly select 10 classes as our leave-one-class-out classes. Additionally, when comparing the performance of OE-based methods, we also use unlabeled datasets as our normal datasets \cite{nips-csi} (e.g., unlabeled CIFAR-10). The details are provided in the \cref{appendix:individual}. 

Throughout all of our experiments on the unimodal and multimodal benchmarks, we train our models using only the training data of the normal class and samples from an OE set that does not include any classes of the anomaly classes in the original classification datasets. We use the area under the receiver operating characteristic curve (AUC) to quantify detection performance. 

\subsection{Comparison Methods}
\noindent
\textbf{End-to-end methods.} We select methods that have demonstrated state-of-the-art performance on the CIFAR-10 and ImageNet-30 unimodal benchmarks. The unsupervised methods we consider include DSVDD \cite{dsvdd}, vanilla-SimCLR \cite{nips-csi}, GT+ \cite{nips-hendrycks}, and CSI \cite{nips-csi}.
For unsupervised OE, we present the results from DSAD \cite{Ruff2020Deep}, HSC \cite{tmlr-oe} and the unsupervised OE variant of GT+ \cite{nips-hendrycks}.
 For supervised OE, we use Focal \cite{focal} and BCE \cite{tmlr-oe}.
 
 \noindent
\textbf{Transfer learning-based methods.} 
We additionally explore transfer learning-based methods. For this category, we select the currently top-performing methods, which include DN2 \cite{dn2}, PANDA \cite{cvpr-panda}, Transformly \cite{transformly}, CLIP \cite{clip} and BCE-CL \cite{tmlr-oe}. All methods are unsupervised with the exception of BCE-CL, which is a fine-tuned version of CLIP with a binary cross-entropy classifier.

The reported results were either extracted from the original papers, if available, or obtained by conducting experiments following the authors's publicly available code (where possible).

\subsection{Results and Analysis} \label{results}
\noindent

\begin{table*}[t]
  \caption{Mean AUC detection performance in $\%$ over 5 trials on the CIFAR-10, ImageNet-30, DIOR and Raabin-WBC in unimodal benchmarks. Bold denotes the best results for training-from-scratch methods, and underline denotes the overall best result. $\ast$ denotes results taken from the reference, $\natural$ denotes results taken from \cite{cvpr-panda}}
  \vskip 0.1in
\resizebox{\textwidth}{!}{
    \begin{tabular}{ccccccccccccc}
\toprule
	& \multicolumn{5}{c}{Unsupervised} & \multicolumn{4}{c}{Unsupervised OE} &\multicolumn{3}{c}{Supervised OE}\\
	\cmidrule(lr){2-6}\cmidrule(lr){7-10}\cmidrule(lr){11-13}
& SimCLR  &  $\text{GT+}$  &  CSI  & self-OECL &CLIP  & $\text{GT+}$ &  $\text{DSAD}$ & HSC & OECL & $\text{Focal}$& BCE & BCE-CL  \\ \midrule
\diagbox[height=2.0em]{Dataset}{From scratch} & $\checkmark$&$\checkmark$&$\checkmark$&$\checkmark$&&$\checkmark$&$\checkmark$&$\checkmark$&$\checkmark$&$\checkmark$&$\checkmark$&  \\ \midrule
CIFAR-10    & $86.1$   & $90.1^\ast$  & $94.3^\ast $  & $94.7 \pm 2.3$& 98.5 & $95.6^\ast$ & $94.5^\ast$ & $95.9^ \ast $ & $\bm{97.8} \pm 0.1$  & $95.8^ \ast$ & $96.1^ \ast $ & $\underline{99.6}^\ast$ \\ 
ImageNet-30 & 64.8  &  $84.8^\ast$       & $91.6^\ast$  & $87.3 \pm 2.1$ & 99.9&$85.7^\ast$ & $96.7^\ast$ & $97.3^ \ast $ & ${\bm{97.7} \pm 0.3}$ & $97.5^ \ast$&${\bm{97.7}}^ \ast$& $\underline{99.9}^\ast$ \\ \midrule
DIOR         &  $72.8 $     & $73.3^\natural$    & $78.5^\natural $  &  $40.4 \pm 6.8$ &87.3&  $79.3$ & $88.3$ & $89.2$    & $\bm{91.1} \pm 0.7$ &$86.7$&88.5 & $\underline{97.7}$ \\ 
Raabin-WBC         & 85.9    &   $73.6$   &  $62.3$  &  $62.1 \pm 5.4$ & 50.0& $77.7$ & $69.3$ & $60.6$ &  $\underline{\bm{88.6}}\pm 1.9$   & 68.9 &66.5 & 51.0\\  
\bottomrule
\end{tabular}%
}
\label{tab:1}%
\vskip 0.1in
\end{table*}%

\begin{table*}[t]
  \caption{Mean AUC detection performance in $\%$ over 2 trials on the CIFAR-10, ImageNet-30, DIOR and Raabin-WBC in multimodal benchmarks. Bold denotes the best results for training-from-scratch methods, and underline denotes the overall best result. $\ast$ denotes results taken from the reference, $\natural$ denotes results taken from \cite{transformly}.}
 \vskip 0.1in
\resizebox{\textwidth}{!}{
    \begin{tabular}{ccccccccccc}
\toprule
	& \multicolumn{6}{c}{Unsupervised} & \multicolumn{2}{c}{Unsupervised OE} &\multicolumn{2}{c}{Supervised OE}\\
	\cmidrule(lr){2-7}\cmidrule(lr){8-9}\cmidrule(lr){10-11} 
& DSVDD  &  SimCLR  &  $\text{DN2}^\ast$  & $\text{PANDA}^\ast$  & $\text{Transformly}^\ast$ & CLIP &  HSC & OECL & BCE & $\text{BCE-CL}$\\ \midrule
\diagbox[height=2.0em]{Dataset}{From scratch} & $\checkmark$ & $\checkmark$ &&&&& $\checkmark$& $\checkmark$&$\checkmark$ & \\ \midrule
CIFAR-10 & $50.7^\natural$ & $83.9$ & $71.7$ & $78.5$ & $90.4$ & 92.2 & $84.8^\ast$ & $\bm{94.6} \pm 0.6$ & $86.6^\ast$ & $\underline{98.4}^\ast$ \\
ImageNet-30 & $\times$ & $94.8$ & $\times$ & $\times$ & $\times$ & 97.7 &$86.7^\ast$ & $\bm{98.8} \pm 0.1$ & $86.5^\ast$ & $\underline{99.2}^\ast$ \\ \midrule
DIOR & $56.7^\natural$ & $\underline{\bm{87.7}}$& $81.1$ & $86.9$ & $66.7$& 66.0& $55.5$ & $87.3  \pm 1.9$ & $62.5$ & $61.4$ \\
Raabin-WBC & $\times$ & $\underline{\bm{83.6}}$ & $\times$ & $\times$ & $\times$ & 54.4 & $61.4$ & $82.2\pm 3.3$ & $62.1$ & $61.3$ \\
\bottomrule
\end{tabular}%
}
\label{tab:2}%
\end{table*}%
\noindent
Our main results are provided in \Cref{tab:1} and \Cref{tab:2}. Overall, our proposed methods achieve performance comparable to the current state-of-the-art methods and mostly outperform training-from-scratch approaches. A particularly surprising outcome is that our self-OECL surpasses CSI in the CIFAR-10 one vs. rest benchmark, while also obtaining decent performance in ImageNet-30. These results support our principles that enhancing the separation of normal and OOD samples in the contrastive feature space is beneficial for OOD detection. Furthermore, this property holds promising potential for future advancements in fully unsupervised OOD detection methods, particularly when more sophisticated OOD generation techniques are introduced (e.g., SDE-based methods \cite{mirzaei2022fake}), transcending the limitations of rotations.

\noindent
\textbf{The benefits of OE datasets.} Our OECL significantly enhances OOD detection compared to the baseline SimCLR in nearly all benchmarks, showcasing substantial improvements. The benefits of ``anomalousness" from OE are especially evident for the unimodal non-standard datasets. Notably, the performance of CSI, a state-of-the-art non-OE method, as expected, shows minimal improvement and even degradation, with AUC scores of $78.5 \%$ for DIOR and $62.3 \%$ for Raabin-WBC, respectively. This is due to the fact that applying rotations no longer shifts the distribution of symmetry image datasets like DIOR and Raabin-WBC, which is the fundamental assumption of CSI. In contrast, the shift-distribution free OECL achieves higher scores of $91.1 \% (+12.6\%)$ and $88.6 \% ( + 26.3\%)$ for these datasets, which indicates the effectiveness of incorporating OE when working with non-standard datasets.

\noindent
\textbf{The power of contrastive learning.} As reported in \cite{tmlr-oe} and further supported by our experiments, there are situations where the current OE techniques struggle, such as the leave-one-class-out benchmark or non-standard training datasets. Particularly in the leave-one-class-out benchmark, where the training dataset's distribution is multimodal, the strategy of encouraging the concentration of normal representations used in HSC \cite{tmlr-oe} is ineffective, leading to suboptimal performance. Our proposed OECL, on the other hand, outperforms the majority of unsupervised methods in the multimodal setting, including those that used pre-trained features, as shown in \Cref{tab:2}. Notably, OECL achieves a remarkable AUC score of $98.8 \%$ for the rich and diversified ImageNet-30, falling behind OE transfer learning-based BCE-CL by only $0.4\%$. These outcomes imply that, in contrast to HSC and BCE, the ability of OECL to exploit high-quality features through contrastive learning is especially advantageous in the context of multimodal datasets, particularly those that are rich and diverse standard datasets. 

\noindent
\textbf{The necessity of training-from-scratch models.} Suprisingly, the results on Raabin-WBC and multimodal DIOR show significant drops in performance for BCE-CL compared to our proposed OECL or SimCLR. This can be attributed to the fact that OECL is trained from scratch, whereas BCE-CL is fine-tuned over pre-training CLIP \cite{clip}, which might not see many examples of $\left\langle \text{text}, \text{image} \right\rangle$ pairs for medical images along its training process. These results highlight the necessity of train-from-scratch models and raise concerns for the generalization of transfer learning in AD to specialize datasets.

\noindent

\begin{figure}[t]
\vskip -0.2in
     \centering
     \includegraphics[width=0.5\textwidth, page=1]{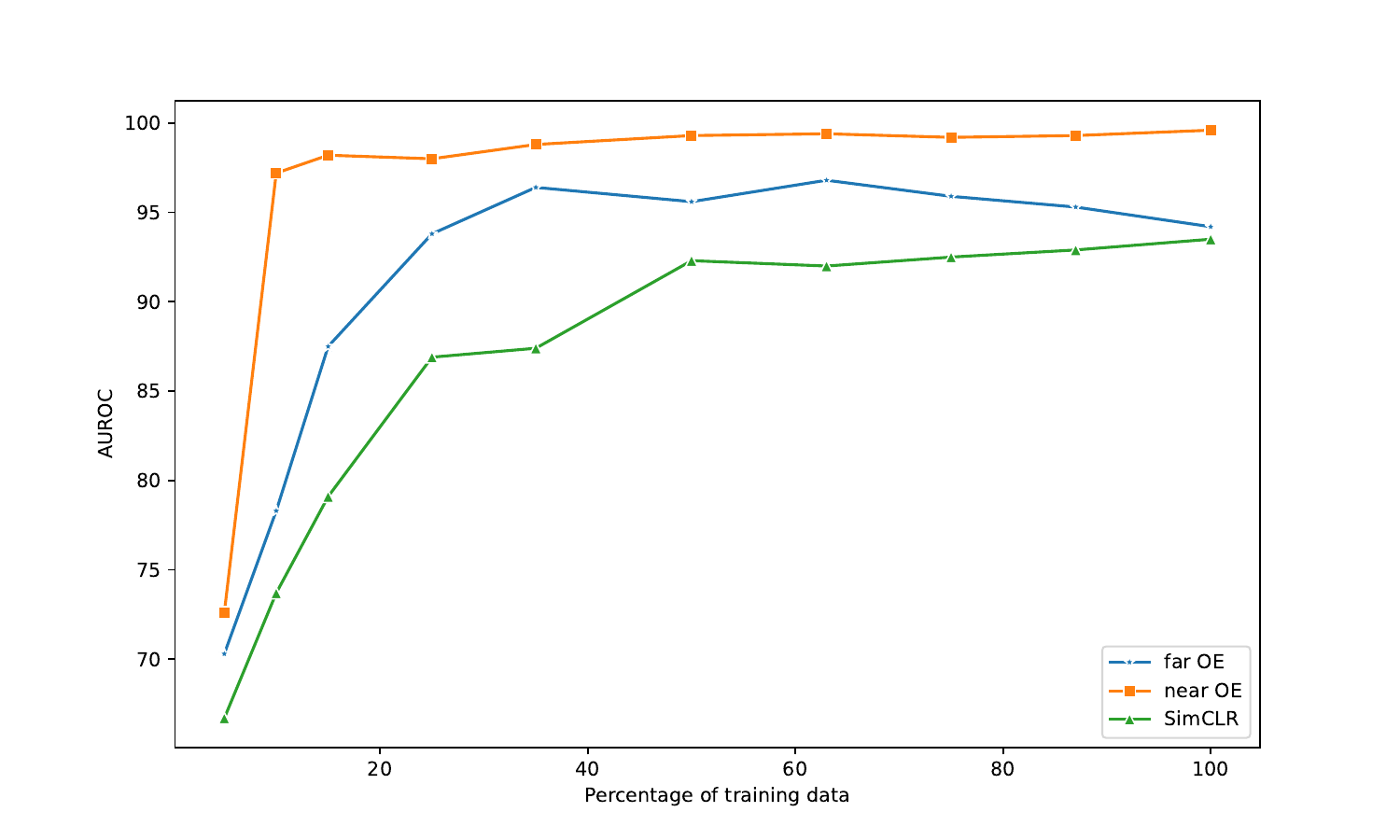}
     \caption{AUROC scores across varying training data sizes for ``near" and ``far" OE datasets. The training dataset consists of lymphocyte and monocyte images, with eosinophil images from the Raabin-WBC dataset employed for testing. A combination of basophil and neutrophil images serves as the ``near" OE, while ImageNet images serve as the ``far" OE. As the training size increases, the influence of the ``far" OE on anomaly detection diminishes, while ``near" OE continues to synergize with the training data.}
        \label{fig:new2}
\vskip 0.2in
\end{figure}
\noindent
\textbf{Diminishing effect and settings where OE are less informative.}   While it's evident that OE datasets significantly contribute in improving anomaly detection in most of our experiments with OECL, their impact is minimal in the settings involving the multimodal non-standard datasets. This discrepancy may be attributed to the quality of deep features learned through $\mathcal{L}_{\text{contrastive}}$ and the relationship between OE datasets and training datasets. Empirically, we observe that as the quality of learned features improves, OE datasets that are significantly different from the training distribution tend to have diminishing effects on AD performance, whereas OE datasets closer to the training distribution continue to enhance AD accuracy. We term this phenomenon the \emph{diminishing effect} of OECL.

To demonstrate this, we conduct experiments using OECL with both ``far" and ``near" OE datasets on Raabin-WBC while gradually reducing the size of our training datasets. Given the difficulty in acquiring ``near" OE data for Raabin-WBC, we designate subclasses basophil and neutrophil as ``near" OE. We then train OECL on a combination of lymphocyte and monocyte images as normal data, with eosinophil images as testing data. For the ``far" OE dataset, we again select ImageNet. The results in \cref{fig:new2} suggest that, as the quality of learned features from contrastive learning improves, the contribution of the ``far" OE to anomaly detection diminishes, even though it significantly boosts anomaly detection performance when the learned features are of lower quality. In contrast, the features from the ``near" OE dataset keep providing useful information to the model.

We also observed a more severe negative effect of ``far" OE on the OE-based BCE-CL, which is a fine-tuned version of the sophisticated features-extractor CLIP \cite{clip, tmlr-oe}. Using unlabeled CIFAR-10 as the normal dataset and CIFAR-100 as OOD images, we measures BCE-CL performances with OE datasets as 80MTI, SVHN \cite{svhn}, and DTD \cite{dtd}, respectively. Our results on \cref{tab:new4} reveal that BCE-CL performances with 80MTI as OE easily achieves state-of-the-art results with an AUROC of $95.9\%$, a notable improvement over the the baseline zero-shot CLIP with an AUROC of $90.8\%$. Conversely, when SVHN and DTD are used as OE datasets, the AUROC scores degrade to $62.8\%$ and $60.5\%$, respectively. On the other hand, our proposed OECL achieves AUROC scores of $94.3\%$ for 80MTI, $89.9\%$ for SVHN, and $90.3\%$ for DTD as OE datasets, significantly improving from the baseline SimCLR with an AUROC of $77.6\%$. These findings underscore the vulnerability of the state-of-the-art OE-based BCE-CL to ``far" OE datasets, while demonstrating the robustness of OECL. Furthermore, these findings complement previous works \cite{tmlr-oe}, which suggest that the inclusion of OE transforms the anomaly detection problem into ``a typical supervised classification problem that does not require a compact decision boundary". Explicitly, the inclusion of OE requires a ``sufficiently" compact decision boundary to achieve non-trivial AD performance.

\begin{table}[ht!]
  \caption{AUC detection performance for BCE-CL and OECL with unlabeled CIFAR-10 as training dataset and CIFAR-100 as testing dataset.}
    \begin{tabular}{cc|ccccc}
\toprule
Dataset & OE Dataset &CLIP & BCE-CL & SimCLR& OECL \\ \midrule
\multirow{3}{*}{\makecell{Unlabeled\\ CIFAR-10}} & 80MTI &\multirow{3}{*}{90.8} & ${95.9}$& \multirow{3}{*}{77.6}  &94.3\\
& SVHN & &${62.8}$  && 89.9 \\
& DTD & &${60.5}$  && 90.3\\

\bottomrule
\end{tabular}%
\label{tab:new4}%
\end{table}%

Although our OECL proved its robustness, the above observations definitively highlight the necessity of carefully applying OE-based methods in real-life scenarios. More experiments and discussions over these phenomenons are given in \cref{appendix:diminish} and \cref{appendix:ovr}.

\noindent
\section{Applicability of OECL} \label{applicability}
\noindent
\subsection{Where and when to apply OECL} \label{OECL:fails}
\begin{figure}[t!]
\vskip -0.2in
     \centering
     \includegraphics[width=0.5\textwidth, page=1]{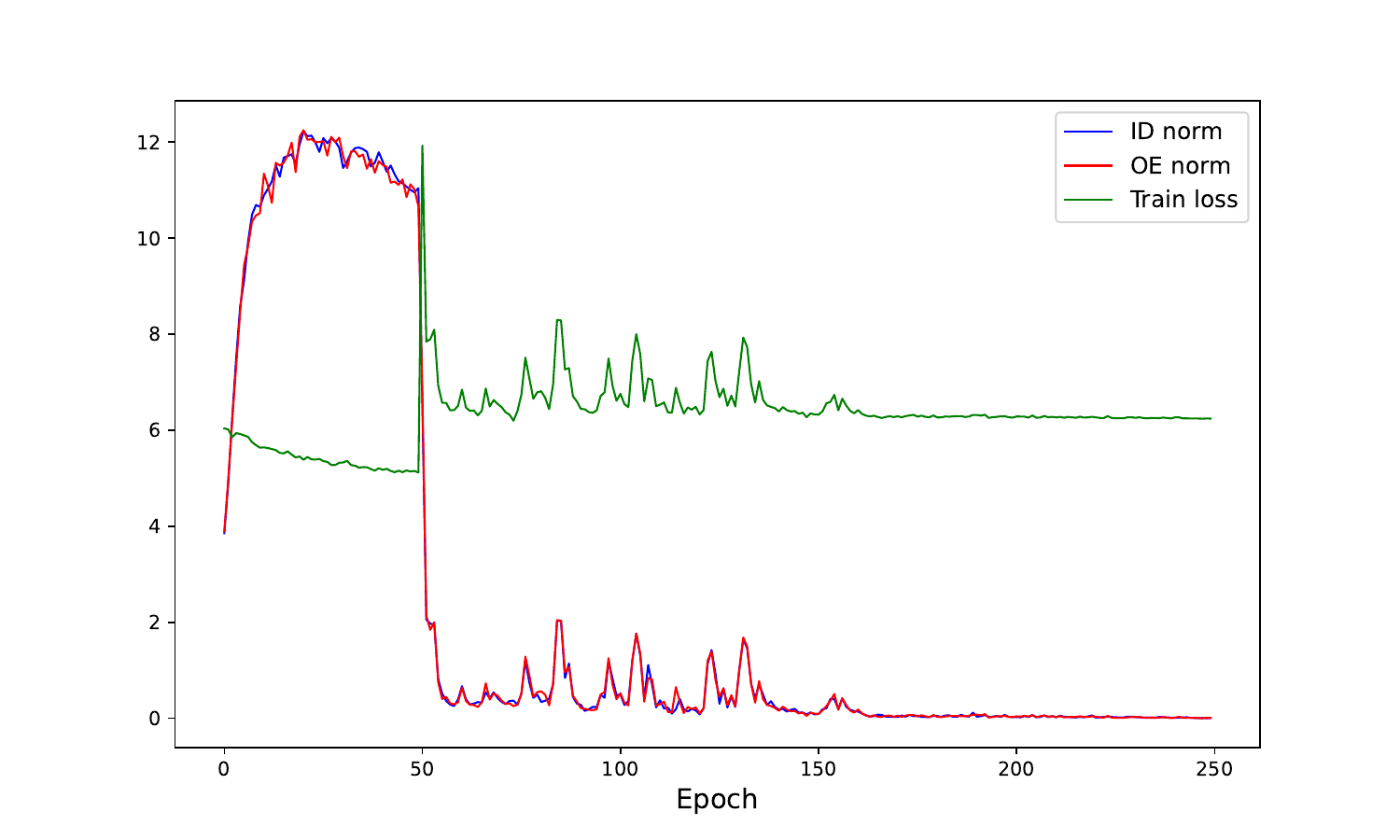}
     \caption{The training loss and $\ell_2$-norm of ID and OE data when we train self-OECL on class monocyte of Raabin-WBC. When OE data and training data are overlap, directly minimizing $\ell_2$-norm of contrastive features makes the training becomes unstable and learns meaningless features (during the initial 50 epochs, we set $\alpha=0$, more information about training is presented in \cref{appendix:warmup}).}
        \label{fig:new1}
\vskip 0.2in
\end{figure}

\noindent
\textbf{Scenarios where OECL fails.} The results of self-OECL on two rotation-invariant datasets Raabin-WBC and DIOR point out an obvious vulnerability of our approach. When there's substantial overlap between the training datasets and the OE datasets, directly minimizing the $\ell_2$-norm of contrastive features make our training process unstable, as shown in \cref{fig:new1}, leading to the learning of trivial features. This observation emphasizes the importance of carefully considering the dataset characteristics when training OECL in practice.

Furthermore, another drawback of OECL is small training datasets. Given that OECL relies on contrastive representation learning to extract high-quality features, it struggles when confronted with datasets where $\mathcal{L}_{\text{contrastive}}$ module is unable to learn useful features and OE is not informative. OECL performances on small datasets such as MVTec-AD significantly lags behind specialized methods \cite{efficientad, recall} by a huge margin ($+ 30\%$). Detailed experiments on small datasets are provided in \cref{appendix:smalldatasets}.

\noindent
\textbf{OECL as a robust, powerful and flexible tool for AD.} OECL had prove its strength and robustness in a variety of settings and datasets. As a training-from-scratch model, OECL is viable in the cases where pre-trained network cannot be used. For example, when one must train a network from scratch due to architectural considerations (e.g., hardware constraints), or when there are security concerns regarding the white-box nature of pre-trained networks \cite{whitebox}. Additionally, the generalization of pre-trained models to specialize data, such as medical images, remains uncertain, underscoring the importance of training-from-scratch models in practical applications, as evidenced in \cref{tab:1} and \cref{tab:2}.

\begin{table}[t]
  \caption{Mean AUC detection performance in $\%$ of OECL over 5 trials on the Raabin-WBC and HAM1000 few-shot OOD benchmarks. Bold denotes the best results.}
   \resizebox{0.48\textwidth}{!}{\begin{tabular}{cccccccc}
\toprule
\multirow{2}{*}{Normal Class} &  \multirow{2}{*}{OE} &\multirow{2}{*}{Method} & \multicolumn{4}{c}{\# Samples} \\
\cmidrule(lr){4-7}& & &k=0 & k=1 & k=10 &\text{Full-shot} \\ \midrule

\multirow{4}{*}{Neutrophils} & \multirow{4}{*}{Lymphocytes}  & OECL &$\bm{68.1}$ &${76.9}$ & ${79.9}$ &83.6\\ 
&& BCE & $\times$&49.0 & 72.6 & ${95.0}$ \\
&& HSC & $\times$&64.7 & 69.1 & 86.1 \\
&& BCE-CL & 62.8 & $\bm{61.8}$ & $\bm{86.4}$ & $\bm{96.1}$\\ \midrule

\multirow{4}{*}{\makecell{Melanocytic\\ nevi}} & \multirow{4}{*}{Melanoma} &OECL& $\bm{63.2}$ & $\bm{65.6}$ &$\bm{69.4}$ & $\bm{83.9}$ \\
 &&BCE& $\times$ & 61.7 &60.4 & 79.1 \\
 &&HSC& $\times$ & 64.1 &46.0 & 75.6 \\ 
 && BCE-CL & 50.9 & 64.0 & 68.4 & 79.8\\ \midrule
 
  \multirow{4}{*}{\makecell{Melanocytic\\ nevi}}&\multirow{4}{*}{\makecell{Benign\\ keratosis}}& OECL& $\bm{59.3}$ & 61.6 & $69.4$ & 85.7\\
  &&BCE &$\times$&53.2 &54.3 & 84.8\\
  &&HSC &$\times$& $\bm{64.7}$ &69.1 & $\bm{88.0}$ \\
  &&BCE-CL & $51.5$ & 50.78& $\bm{76.2}$ & 85.3
  \\ \bottomrule

\end{tabular}}
\label{tab:4}
\end{table}

\subsection{Few-shot OOD detection}
\noindent
We explore the generalization of OECL to unseen anomalies. In many real-life scenarios, we are able to collect some samples of a few specific classes of anomalies, which we term as ``obtainable OOD classes". For instance, in hematology, suppose that neutrophils, representing about $60\%$ of circulating white blood cells, are regarded as normal samples, while white blood cells of other types are considered anomalies. Within this context, lymphocytes, which make up about $30\%$ of white blood cells, can be considered an obtainable OOD class, and other white blood cells like basophils ($\approx 1\%$) and eosinophils ($\approx 3\%$) represent ``unseen anomalies".

\noindent
\textbf{Few-shot OOD benchmark.} Inspired by \cite{iclr-ssd} and \cite{nips-step}, to evaluate OOD detection performance against unseen anomalies described above, we introduce the ``few-shot OOD'' benchmark. This benchmark is constructed using classification datasets (e.g., Raabin-WBC), where the class with the largest number of samples (e.g., neutrophils) is considered as normal. The next largest class (e.g., lymphocytes) is selected as the obtainable OOD class, for which a small number of samples are available during training. All other classes (e.g., basophils, eosinophils, and monocytes) serve as unseen anomalies. At inference time, the goal is to identify normal samples and unseen abnormal samples. In this study, we focus on the extreme case one-shot and ten-shot detection, implying access to only a single sample and ten samples of obtainable OOD datasets, respectively \footnote[1]{To some degree, the few-shot OOD benchmark represents unimodal benchmarks with very few standard OE samples.}.

\noindent
\textbf{Datasets.} In addition to the Raabin-WBC dataset, we use the HAM10000 dataset \cite{ham10000}, which is a collection of dermatoscopic images of common pigmented skin lesions. Since the number of samples in the second and the third largest classes of HAM10000 are similar, we choose to do two experiments, where each of them serve as the obtainable OOD class.

\noindent
\textbf{Results and discussion.} Results from \Cref{tab:4} show that exposing OECL to a small number of samples from a specific near-OOD class helps improve anomaly detection accuracy significantly. For less diverse Raabin-WBC, only one single OOD sample helps to boost the AUC score up about $9\%$. These results suggest a practical solution in which a limited number of OOD samples \cite{li2021cutpaste, Zavrtanik_2021_ICCV} are integrated into the training phase to improve AD performance.

\section{Conclusion} \label{conclusion}
\noindent
We propose an easy-to-use but effective technique named outlier exposure contrastive learning, which extends the strengths of both contrastive learning and OE for out-of-distribution detection problems. OECL demonstrates strong and robust performance under various OOD detection scenarios. Additionally, we investigate scenarios where OE datasets might degrade the anomaly detection performance of OE-based methods. We believe that our work will serve as an essential baseline for future outlier exposure approaches to anomaly detection.

\section*{Broader impact}
\noindent
This study focuses on the domain of \emph{out-of-distribution} (OOD) detection, which is a crucial part of trustworthy intelligent systems. Our research offers both theoretical and empirical findings regarding the behavior of $\ell_2$-norm in contrastive learning. Additionally, we propose a simple technique for directly incorporating OOD samples into contrastive learning methods. By leveraging ``anomalousness'' of OOD samples and high-quality representations of contrastive learning, our work will provide a feasible solution to problems where current OE and contrastive AD methods fail. We expect that these results will have an impact on real-life applications as well as academic advancements, leading to the development of safer and more dependable AI systems.

\bibliographystyle{abbrvnat}
\bibliography{ref.bib}

\onecolumn
\appendix

\subsection{Proof of \Cref{thm:1}}

\begin{lemma} \label{lemma:1}
For any $\mu>0$, $f(y)$ is a non-increasing and convex function on $\left[0, \infty\right)$ where
\begin{align*}
f(y) = \int_{-\infty}^{\infty} \frac{x}{\sqrt{x^2+y}} e^{-(x-\mu)^2/2\sigma^2}dx
\end{align*}
\end{lemma}

\begin{proof}
Firstly, rewrite
\begin{align*}
f(y) &= \int_{0}^{\infty} +\int_{-\infty}^{0} \frac{x}{\sqrt{x^2+y}} e^{-(x-\mu)^2/2\sigma^2}dx \\
&= \int_{0}^{\infty} \frac{x}{\sqrt{x^2+y}} \left(e^{-(x-\mu)^2/2\sigma^2} - e^{-(x+\mu)^2/2\sigma^2} \right)dx\\
&= \int_{0}^{\infty} \frac{x}{\sqrt{x^2+y}} e^{-(x+\mu)^2/2\sigma^2} \left(e^{2x\mu/\sigma^2} - 1\right)dx
\end{align*}
$f'(y)$ and $f''(y)$ given by
\begin{align*}
f'(y) &= \left(\int_{0}^{\infty} \frac{x}{\sqrt{x^2+y}} e^{-(x+\mu)^2/2\sigma^2} \left(e^{2x\mu/\sigma^2} - 1\right)dx \right)_y' \\
&= \int_{0}^{\infty} \left(\frac{x}{\sqrt{x^2+y}}\right)_y' e^{-(x+\mu)^2/2\sigma^2} \left(e^{2x\mu/\sigma^2} - 1\right)dx \\
& = \int_{0}^{\infty} \frac{-x}{2(x^2+y)^{3/2}} e^{-(x+\mu)^2/2\sigma^2} \left(e^{2x\mu/\sigma^2} - 1\right)dx\\
f''(y) &= \left(\int_{0}^{\infty} \frac{x}{\sqrt{x^2+y}} e^{-(x+\mu)^2/2\sigma^2} \left(e^{2x\mu/\sigma^2} - 1\right)dx \right)_y'' \\
&= \int_{0}^{\infty} \left(\frac{x}{\sqrt{x^2+y}}\right)_y'' e^{-(x+\mu)^2/2\sigma^2} \left(e^{2x\mu/\sigma^2} - 1\right)dx \\
& = \int_{0}^{\infty} \frac{3x}{(x^2+y)^{5/2}} e^{-(x+\mu)^2/2\sigma^2} \left(e^{2x\mu/\sigma^2} - 1\right)dx
\end{align*}

Since $e^{2x\mu/\sigma^2} \geq 1 ~\forall x>0$, it's clear that $f'(y)<0$ and $f''(y)>0$, i.e, $f(y)$ is non-increase and convex over $\left[0, \infty \right)$.
\end{proof}

\printProofs

\subsection{Gradual reduction of the $\ell_2$-norm of contrastive features} \label{appendix:red}

In this section, to study the effect of \emph{alignment} and \emph{uniformity} on the $\ell_2$-norm of contrastive features, we use the contrastive learning framework proposed by \cite{icml-wang},
$$\mathcal{L}_{\text{contrastive}} = \mathcal{L}_{\text{align}} + \mathcal{L}_{\text{uniform}},$$
where
\begin{align}
\mathcal{L}_{\text{align}} &= \underset{(x, y) \sim p_{\text{pos}}}{\mathbb{E}} \left[\Vert z(x) - z(y)\Vert_2^2\right],\\ 
\mathcal{L}_{\text{uniform}} &= \log  \underset{\substack{(x, y) \overset{\text{i.i.d}}{\sim} p_{\text{data}}}}{\mathbb{E}} \left[e^{-\Vert z(x) - z(y)\Vert_2^2}\right].
\end{align}

\begin{figure}
     \centering
       \includegraphics[width=0.8\textwidth, page=1]{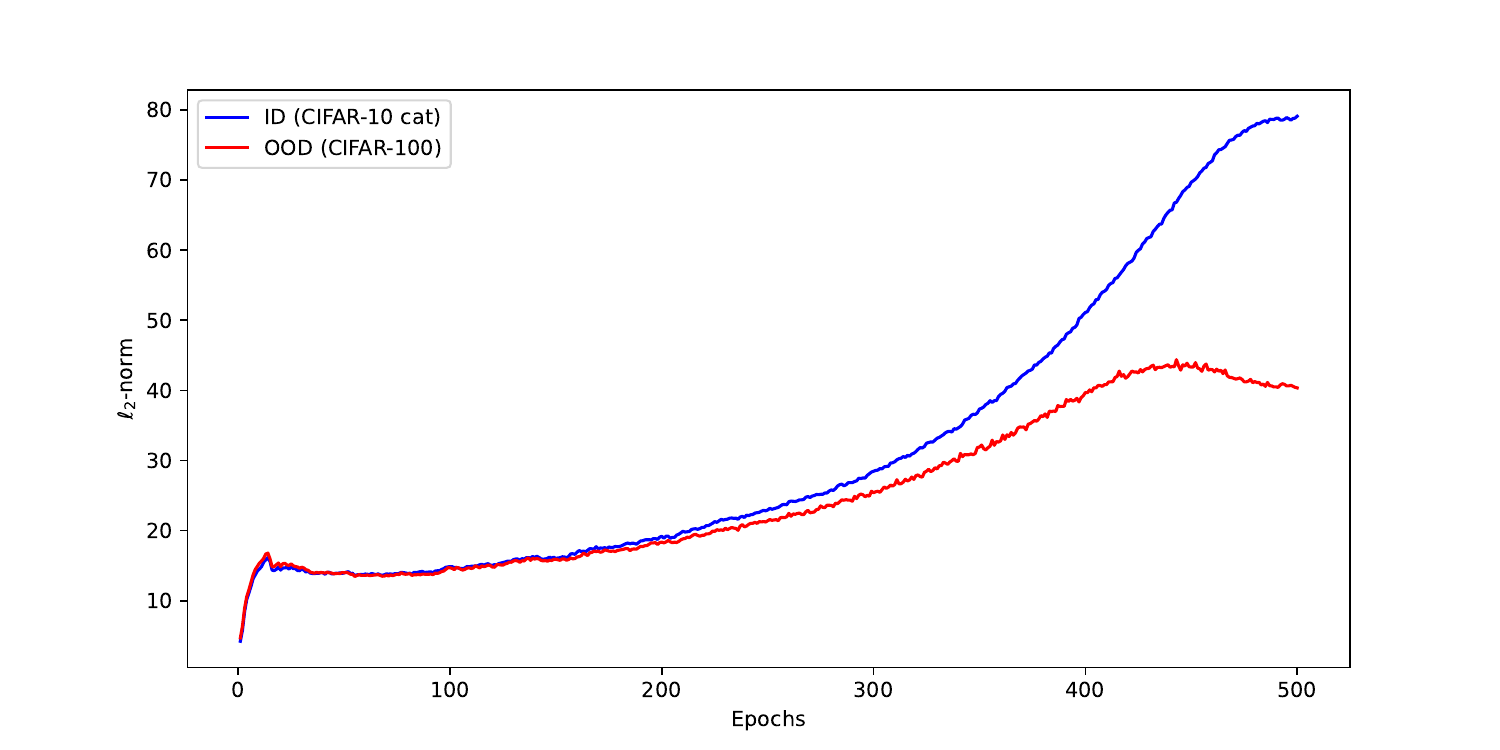}
        \caption{ $\ell_2$-norm of contrastive features for both normal and OOD samples during optimization of $\mathcal{L}_{\text{contrastive}}$ without $\ell_2$-normalization, using only normal samples.}
        \label{fig:5}
\end{figure}

\noindent
\textbf{Necessity of cosine similarity.}~\Cref{fig:5} depicts the $\ell_2$-norm of contrastive features during optimization of $\mathcal{L}_{\text{contrastive}}$ using Euclidean distance, i.e., $z(x) = f(x)$. As expected, without $\ell_2$-normalization, the $\ell_2$-norm of contrastive features $f(x)$ does not exhibit the characteristic of gradual decrease but rather increases as training progresses. This shows the critical role played by cosine similarity in driving the gradual reduction of the $\ell_2$-norm of contrastive features.

\begin{figure}[h]
\centering
     \subfigure[$\ell_2$-norm of contrastive features.]{\includegraphics[width=8cm, height=5cm, page=1]{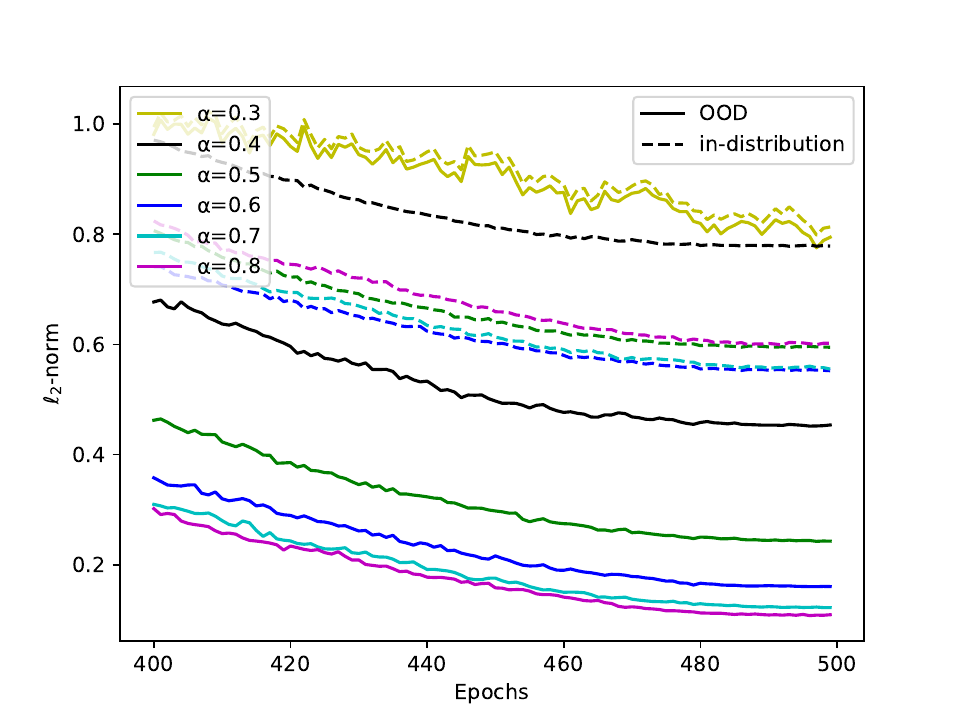}}
     \hfill
    \subfigure[$\mathcal{L}_{\text{align}}, \mathcal{L}_{\text{uniform}}$ and AUC score.]{\includegraphics[width=8cm, height=5cm, page=1]{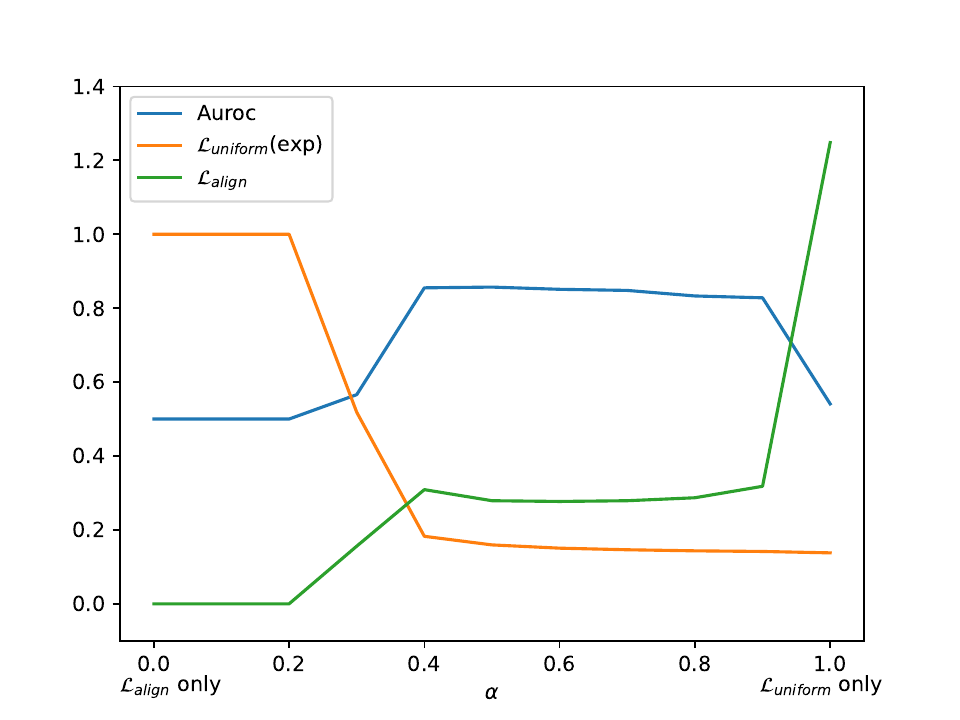}}     
        \caption{Impact of optimizing different weighted combinations of $\mathcal{L}_{\text{align}}$ and $\mathcal{L}_{\text{uniform}}$, trained on CIFAR-10 cat class. $\mathcal{L}_{\text{uniform}}$ is exponentiated for visualization.}
        \label{fig:6}
\end{figure}

\newpage

\noindent
\textbf{Necessity of both alignment and uniformity.} \Cref{fig:6} shows how the AUC score and the $\ell_2$-norm change in response to optimizing differently weighted combinations  $(1-\alpha)\mathcal{L}_{\text{align}}+ \alpha\mathcal{L}_{\text{uniform}}$. Similar to the accuracy curve in \cite{icml-wang}, AUC score curve of OOD detection is also inverted-U-shaped. This means that both \emph{alignment} and \emph{uniformity} are vital for a high-quality features extractor, hence is necessity for a good OOD detector. When $\mathcal{L}_{\text{align}}$ is weighted much higher than $\mathcal{L}_{\text{uniform}}$, degenerate solution occurs and all inputs (including normal and OOD samples) are mapped to the same point ($\exp(\mathcal{L}_{\text{uniform}})$=1). However, as long as the ratio between two weights is not too large (e.g., $0.4 \leq \alpha \leq 0.9$), we consistently observe both the gap between the $\ell_2$-norm of normal and OOD contrastive representations and their gradual reduction as training progresses. An intriguing observation is the relation between $\alpha$ and the $\ell_2$-norm of OOD samples; as the value of $\alpha$ increases, the $\ell_2$-norm of OOD contrastive features decreases proportionally. This suggests that uniformity plays a more significant role in the gradual reduction of $\ell_2$-norm of contrastive features.  We hope that future research will offer a more in-depth understanding of the fundamental cause of this phenomenon.

\subsection{Anomaly detection scores} \label{appendix:score}

We consider the following anomaly detection scores,
\begin{align}
s_{\ell_2} &= \lVert f(x) \rVert_2^2\\
s_{\ell_2-\text{ens}} &=  \mathbb{E}_{t\in T}\lVert f(t(x)) \rVert_2^2 = \mu^2 + \sigma^2 + \sigma_V^2\\
s_{\mu} &= \lVert\mathbb{E}_{t\in T}\lVert f(t(x)) \rVert_2^2 =  \mu^2
\end{align}
Experimentally, we find that the overall detection performance of $s_{\mu}$ is better, especially for non-standard datasets as shown in \cref{tab:5}.

\begin{table}[ht!]
  \caption{AUC detection performance in $\%$ using $s_{\ell_2}$, $s_{\ell_2-\text{ens}}$ and $s_\mu$.}
    \begin{tabular}{cc|ccc}
\toprule
Dataset &  Benchmark & $s_\mu$ & $s_{\ell_2-\text{ens}}$ & $s_{\ell_2}$ \\ \midrule
 \multirow{2}{*}{DIOR} & unimodal &$\bm{91.1}$ & 90.7 & 90.8 \\
 & multimodal & $\bm{87.3}$ & 86.9 &87.1\\ \midrule 
  \multirow{2}{*}{Raabin-WBC} & unimodal & $\bm{88.6}$ & 88.4 & $\bm{88.6}$\\
  & multimodal & $\bm{82.2}$ & 82.0 & 81.4\\ \midrule 
    \multirow{2}{*}{CIFAR-10} & unimodal & $\bm{97.8}$ & 97.5 & $96.4$\\
  & multimodal & $\bm{94.6}$ & 94.4 & 93.8\\ \midrule 
      \multirow{2}{*}{ImageNet-30} & unimodal & $\bm{97.7}$ & $\bm{97.7}$ & $95.9$\\
  & multimodal & $\bm{98.8}$ & $\bm{98.8}$ & 97.8\\

\bottomrule
\end{tabular}%
\label{tab:5}%
\end{table}%
Another potential addition to the anomaly scores involves incorporating the cosine similarity to the nearest sample as suggested in \cite{nips-csi}. While we have observed a similar phenomenon in our experiments, where the combination of cosine similarity and feature norm mostly yields the best results, it may not exhibit absolute consistency across various scenarios, as indicated in \Cref{tab:6}. Furthermore, we wish to emphasize the effectiveness of the $\ell_2$-norm in anomaly detection. Therefore, we have chosen to present our anomaly score without the inclusion of cosine similarity.
\begin{table}[ht!]
  \caption{AUC detection performance in $\%$ using $s_\mu$ and $\text{sim} + s_{\mu}$.}
    \begin{tabular}{ccc|cc}
\toprule
Dataset &  Benchmark & OE&$s_{\mu}$ & $ \text{sim}+s_{\mu}$ \\ \midrule
Raabin-WBC & unimodal& &${85.9}$ & $\bm{88.1}$ \\
Raabin-WBC & unimodal& \checkmark &${88.6}$ & $\bm{89.0}$ \\
DIOR & unimodal &  & $72.8$ & $\bm{74.9}$\\
DIOR & unimodal & \checkmark & $91.1$ & $\bm{91.2}$\\
CIFAR-10 & unimodal &  &${86.1}$ & $\bm{86.8}$\\
CIFAR-10 & unimodal & \checkmark  &$\bm{97.8}$ & ${97.7}$\\
ImageNet-30 & unimodal & &${64.8}$ & $\bm{66.7}$ \\
ImageNet-30 & unimodal & \checkmark&$\bm{97.7}$ & $97.5$ \\
\bottomrule
\end{tabular}%
\label{tab:6}%
\end{table}%

\subsection{Small datasets}  \label{appendix:smalldatasets}

Due to the fact that OECL relies on contrastive representation learning to extract high-quality features, it does not perform well on datasets where its contrastive learning module is unable to learn strong features and OE is not informative. This is particularly evident in scenarios where the training datasets are small and non-standard. To illustrate this, we use the DTD \cite{dtd} one vs. rest benchmark and the recent manufacturing dataset MVTec-AD~\cite{mvtec}, with ImagetNet22k serving as the OE dataset. 

DTD is a dataset that includes different classes of textures. Comprising 15 distinct classes of manufacturing objects (such as screws, cables, and capsules), MVTec-AD presents both normal and abnormal samples for each class. For example, a screw class consists of two sets of images: a training set of images of typical screws, and a test set of images of typical screws, along with images of screws with anomalous defects. Both DTD and MVtec-AD consist of fewer than 40 samples for each normal class. \Cref{tab:7} shows the results of OECL on these small datasets.

\begin{table}[ht!]
  \caption{Mean AUC detection performance in $\%$ over 5 trials on the DTD one vs. rest and on the MVTec-AD benchmarks with ImageNet1k OE.}
    \begin{tabular}{c|ccccc}
\toprule
Dataset & OECL & $\text{HSC}^\ast$ & $\text{BCE}^\ast$ & Patchcore & EfficientAD\\ \midrule
DTD  & 65.6 & 72.7 & ${73.3}$ & $\times$ & $\times$\\
MVTec-AD & 62.1 & ${70.1}$ & 66.1 & $\bm{99.1}$ & $\bm{99.1}$\\ 
\bottomrule
\end{tabular}%
\label{tab:7}%
\end{table}%

As anticipated, OECL results more poorly on the small datasets than state-of-the-art methods: OECL scores around $62 \%$ while \cite{efficientad} and \cite{recall} score above $99\%$ AUC for the MVTec-AD benchmark. These results indicate that for small datasets, OECL may not be the optimal choice.


\subsection{OECL with SupCLR} \label{appendix:supclr} 
In this subsection, we present the OECL version of SupCLR \cite{supclr}. Instead of using SimCLR for $\mathcal{L}_{\text{contrastive}}$ in \cref{loss:1}, we utilize SupCLR. SupCLR is a supervised extension of SimCLR that contrasts samples \emph{class-wise}. Hence, we conduct our experiments using labeled CIFAR-10 with the OE dataset is 80MTI, and the out-of-distribution (OOD) data are common benchmark datasets: CIFAR-100 and SVHN. For comparison, we also include experiments using OECL on unlabeled CIFAR-10. The results show that OE can improve the performance of SupCLR on anomaly detection.

\begin{table}[h!]
  \caption{AUC detection performance of SupCLR and its OE extension on CIFAR-10.}
    \begin{tabular}{c|c|cc}
\toprule
\multirow{2}{*}{Dataset} & \multirow{2}{*}{Method} & \multicolumn{2}{c}{OOD Dataset}\\
\cmidrule(lr){3-4} &&CIFAR-100 & SVHN  \\ \midrule
\multirow{2}{*}{Labeled CIFAR-10} &  SupCLR & 88.6 & 97.3\\
& OE + SupCLR & 95.2 & 97.4\\ \midrule
\multirow{1}{*}{Unlabeled CIFAR-10} &OECL &94.3 & 99.4\\
\bottomrule

\end{tabular}%
\label{tab:new1}%
\end{table}%

\subsection{The diminishing effect.} \label{appendix:diminish}
\begin{figure}[h!]
\vskip -0.2in
     \centering
     \includegraphics[width=0.5\textwidth, page=1]{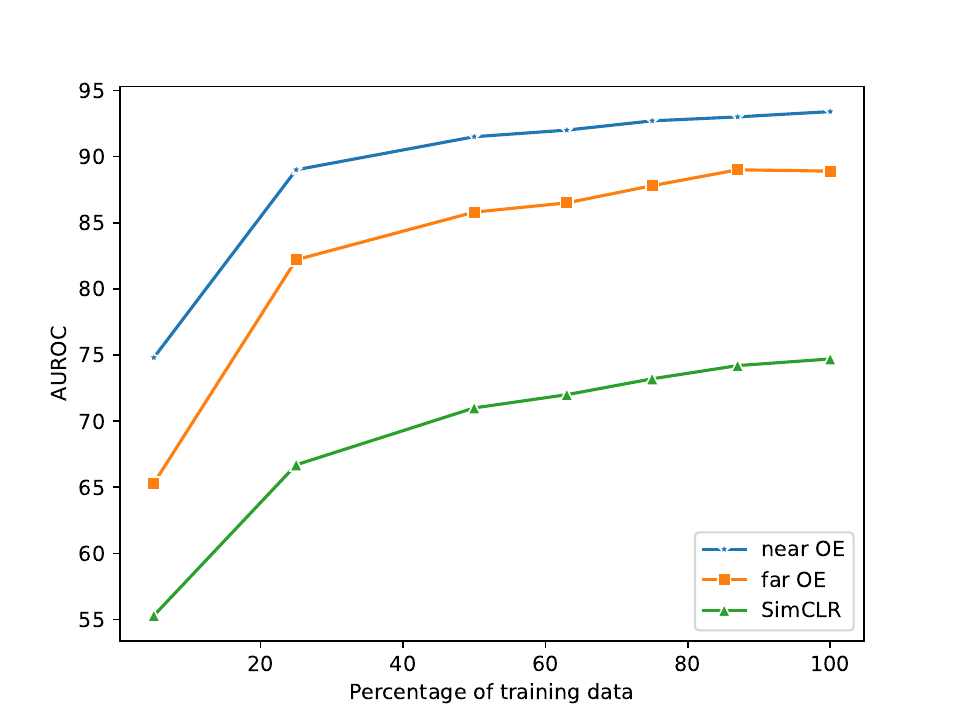}
     \caption{AUROC scores across varying training data sizes for ``near" and ``far" OE datasets. The training dataset is unlabeled CIFAR-10. The test data is CIFAR-100 and we choose SVHN as ``far" OE and keep using 80MTI as ``near" OE dataset.}
        \label{fig:new3}
\vskip 0.2in
\end{figure}

We provide results for unlabelled CIFAR-10 datasets in \cref{fig:new3}. The test data is CIFAR-100 and we choose SVHN as ``far" OE and keep using 80MTI as ``near" OE dataset. 

\subsection{Warm-up with SimCLR} \label{appendix:warmup}
As mentioned in \cref{OECL:fails}, brutally force $\ell_2$-norm of OE samples to zero may cause the instability in training process, especially when OE samples might overlap with training dataset. Since SimCLR uses hard augmentations, for example, CSI \cite{nips-csi} uses RandomResizeCrop with scale uniformly from 0.08 up to 1, augmentations of training samples and OE samples might send wrong signals to the model at the early training steps. For this reason, we decide to warm-up OECL for 50 training epochs with $\mathcal{L}_{\text{contrastive}}$ (i.e., $\alpha=0$ for the first 50 epochs). The following figure presents training process of class deer in CIFAR-10 with and without warm-up.

\begin{figure*}[ht!]
\vskip -0.2in
     \centering
     \subfigure[Warm-up 50 epochs with SimCLR.]{\includegraphics[width=0.48\textwidth, page=1]{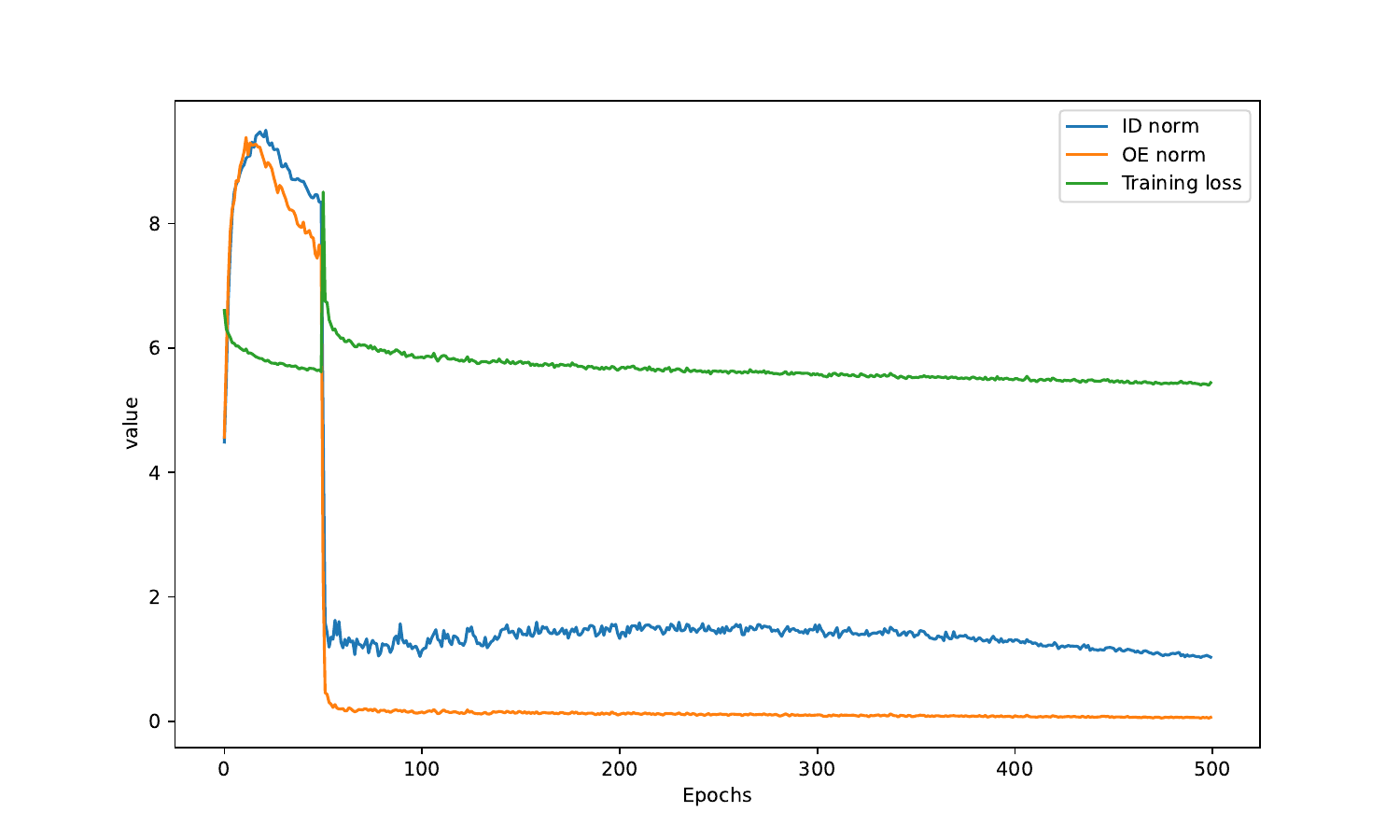}}
     \hfill
     \subfigure[No warm-up.]{\includegraphics[width=0.48\textwidth, page=1]{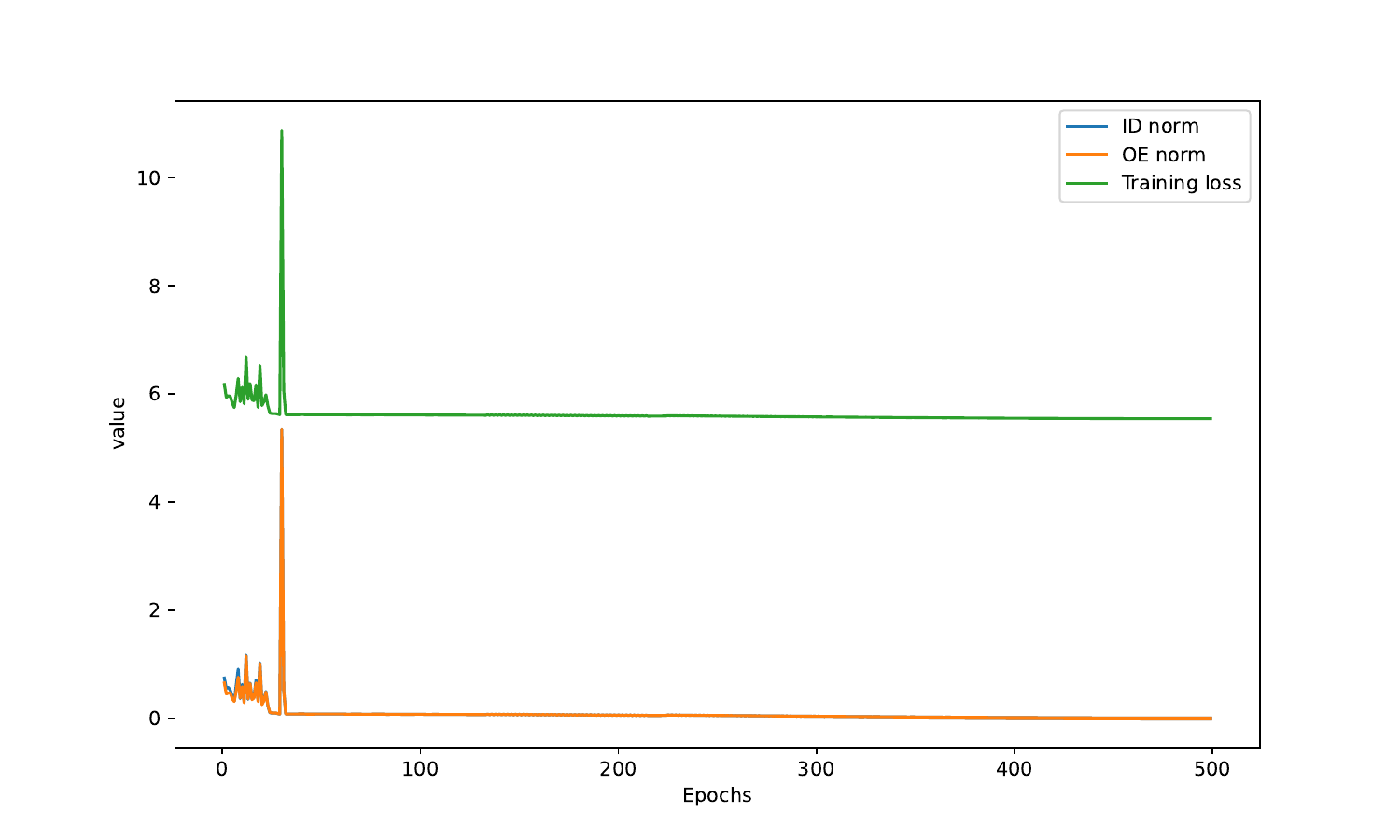}}
        \caption{Training loss and norm of training samples and OE samples during optimization of OECL with and without warm-up.}
        \label{fig:2}
\vskip 0.2in
\end{figure*}

\subsection{Performance of OECL with different OE datasets the one vs. rest benchmark.} \label{appendix:ovr}
To further illustrate the strengths of OECL, we measure the performance of our proposed method under the unimodal setting with SVHN, DTD and one sample from CIFAR-100 as our non-standard OEs. \Cref{tab:3} shows the overwhelming performance of OECL over HSC for non-standard OE datasets. 

\begin{table}[h!]
  \caption{AUC detection performance in $\%$ for OECL and HSC on the one vs. rest CIFAR-10 benchmark.}
    \begin{tabular}{cc|cc}
\toprule
Dataset & OE Dataset & OECL & HSC \\ \midrule
\multirow{3}{*}{CIFAR-10} & SVHN & ${92.2}$ & 54.2 \\
& DTD & ${91.7}$ & 82.9 \\
& 1-sample CIFAR-100 & ${86.9}$ & 50.5\\
\bottomrule
\end{tabular}%
\label{tab:3}%
\end{table}%


\subsection{Implementation details} \label{appendix:exp}
We use Resnet \cite{resnet} along with a 2-layer MLP with batch-normalization and Relu after the first layer. The based contrastive learning module is SimCLR \eqref{loss:simclr} with a temperature of $\tau = 0.5$. We set the balancing hyperparameter $\alpha=1$ for CIFAR-10 and Imagenet-30 datasets and $\alpha=0.1$ for DIOR, Raabin-WBC, and HAM10000 datasets. For optimization, we train all models with 500 epochs except self-OECL with 1000 epochs. We adapt LARS optimizer \cite{lars} with a weight decay of $1e-6$ and momentum of $0.9$. During the first 50 epochs, we set $\alpha = 0$ as a warming-up phase. For the learning rate scheduling, we use learning rate 0.1 and decay with a cosine decay schedule without restart \cite{cosine_decay}. We use the batch size of 256 for both normal and OE samples during training, and we detect OOD samples using the ensemble score function $s_{\mu}$ \eqref{score:1} with 32 augmentations. For results on \cref{fig:new2} and \cref{fig:new3}, we use $s_{\ell_2}$ \eqref{score:0} as our score function. Additionally, we implement global batch normalization \cite{bn}, which shares batch normalization parameters across GPUs in distributed training setups.

\begin{itemize}
\item \textbf{CIFAR-10} includes 10 standard object classes with totally $50000$ images for training and $10000$ images for testing. 
\item \textbf{ImageNet-30} contains $39000$ training and $3000$ testing images that are divided equally into $30$ standard object classes. In leave-one-class-out setting, we randomly selected the following classes: banjo, barn, dumbbell, forklift, nail, parking meter, pillow, schooner, tank, toaster.
\item \textbf{DIOR} is an aerial images dataset with 19 object categories. Following previous papers \cite{transformly}; \cite{cvpr-panda}, we use the bounding box provided with the data, and select classes that have more than 50 images with dimensions greater than $120 \times 120$ pixels. This preprocessing step results in 19 classes, with an average of 649 samples. The class sizes are not equal, with the smallest class (baseballfield) in the training set containing 116 samples and the largest (windmill) containing 1890 samples.
\item \textbf{Raabin-WBC} is a dataset comprising images of white blood cells categorized into five classes. The training set contains more than $10000$ images. For the test set, we choose the Test-A of the dataset \cite{raabin}. 
\item \textbf{HAM10000} is a large collection of dermatoscopic images of common pigmented skin lesions with total $10000$ training images. We follow the instruction given by \cite{ham10000}, using  ISIC2018\textunderscore Task3\textunderscore Test\textunderscore Images.zip (1511 images) for the evaluation purposes.
\end{itemize}
\subsection{Data augmentations}
For data augmentations $T$ and $T_{\text{oe}}$, we use SimCLR augmentations: random crop, horizontal flip, color jitter, and grayscale. We simply set $T_{\text{oe}} = T$ in all our experiments. The details for the augmentations are:
\begin{itemize}
\item \textbf{Random crop.} We randomly crops of the original image with uniform distribution from $a$ to $1.0$. We select $a = 0.5$ for Raabin-WBC dataset and Ham10000 dataset, and all the other datasets we set $a=0.08$. After cropping, cropped image are resized to match the original image size.
\item \textbf{Horizontal flip.} All training images are randomly horizontally flipped with $50\%$ of probability.
\item \textbf{Color jitter.} Color jitter changes the brightness, contrast, saturation and hue of images. We implement color jitter using pytorch's color jitter function with $\text{brightness} = 0.4, \text{contrast} = 0.4, \text{saturation}=0.4$ and $\text{hue}=0.1$. The probability to apply color jitter to a training image is set at $80\%$.
\item \textbf{Grayscale.} Convert an image to grayscale with $20\%$ of probability.
\end{itemize}

\subsection{Implementation Details for Other Evaluated Methods}
\subsubsection{HSC, BCE, Focal, CLIP and BCE-CLIP} We use the official implementations from \cite{tmlr-oe}. We follow the official setup in their paper in all datasets except for \cref{tab:3} and \cref{tab:4}. For the experiment of HSC with 1-sample CIFAR100 as OE, we train it for 200 epochs with milestones at 100 and 150 epochs instead of 30 epochs. So it have same training ground with experiments using SVHN and DTD as OE datasets. For BCE-CLIP configurations on \cref{tab:4}, we train it for 40 epochs with learning rate reduced by a factor of 10 at milestone 35.

\subsubsection{GT and GT+}
We use the official implementations from \cite{nips-hendrycks}. For GT+, we encourage the network to have uniform soft-max responses on OE samples.

\end{document}